\documentclass{article}

\usepackage{microtype}
\usepackage{graphicx}
\usepackage{subfigure}
\usepackage{booktabs} 
\usepackage{amsmath}

\usepackage{hyperref}

\usepackage[accepted]{icml2018}

\icmltitlerunning{Learning Representations and Generative Models for 3D Point Clouds}

\usepackage{multirow}
\usepackage{tabularx}
\newcolumntype{Y}{>{\centering\arraybackslash}X}
\newcolumntype{L}[1]{>{\hsize=#1\hsize\raggedright\arraybackslash}X}%
\newcolumntype{R}[1]{>{\hsize=#1\hsize\raggedleft\arraybackslash}X}%
\newcolumntype{C}[1]{>{\hsize=#1\hsize\centering\arraybackslash}X}%
\usepackage{wrapfig}
\usepackage{bm}
\usepackage{amsfonts}       

\def\vx{{\bm{x}}}

\def\vz{{\bm{z}}}

\newcommand{\textarrowright}{ $\,\to\,$ }

\newcommand{\pdata}{p_{\rm{data}}}

\newcommand{\outline}[1]{}

\renewcommand{\tilde}{{\raise.17ex\hbox{$\scriptstyle\sim$}}}

\begin{document}

\twocolumn[
\icmltitle{Learning Representations and Generative Models for 3D Point Clouds}




\begin{icmlauthorlist}
\icmlauthor{Panos Achlioptas}{su}
\icmlauthor{Olga Diamanti}{su}
\icmlauthor{Ioannis Mitliagkas}{mo}
\icmlauthor{Leonidas Guibas}{su}
\end{icmlauthorlist}

\icmlaffiliation{su}{Department of Computer Science, Stanford University, USA}
\icmlaffiliation{mo}{MILA, Department of Computer Science and Operations Research, University of Montr\'eal, Canada}
\icmlcorrespondingauthor{Panos Achlioptas}{optas@cs.stanford.edu}

\icmlkeywords{pointclouds, GMM, GAN, autoencoder, EMD, matchings}

\vskip 0.3in
]



\printAffiliations{}  

\begin{abstract}
Three-dimensional geometric data offer an excellent domain for studying representation learning and generative modeling. In this paper, we look at geometric data represented as point clouds.
We introduce a deep AutoEncoder (AE) network with state-of-the-art reconstruction quality and generalization ability. The learned representations outperform existing methods on 3D recognition tasks and enable shape editing via simple algebraic manipulations, such as semantic part editing, shape analogies and shape interpolation, as well as shape completion. We perform a thorough study of different generative models including GANs operating on the raw point clouds, significantly improved GANs trained in the fixed latent space of our AEs, and Gaussian Mixture Models (GMMs). To quantitatively evaluate generative models we introduce measures of sample fidelity and diversity based on matchings between sets of point clouds. Interestingly, our evaluation of generalization, fidelity and diversity reveals that GMMs trained in the latent space of our AEs yield the best results overall.
\end{abstract}

\section{Introduction}

\begin{figure*}[t]
    \centering
    \includegraphics[width=\textwidth]{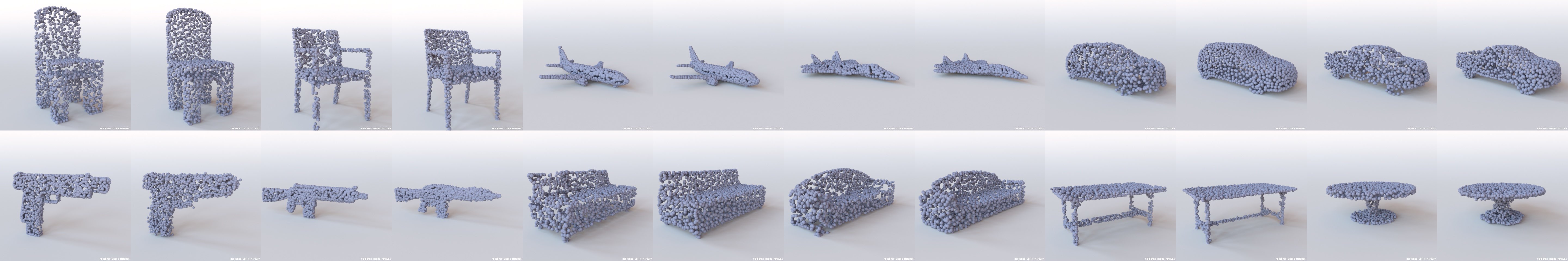}
    \vspace{-20pt}
    \caption{Reconstructions of unseen shapes from the test split of the input data. The leftmost image of each pair shows the ground truth shape, the rightmost the shape produced after encoding and decoding using a {\em class-specific} AE-EMD.}
        \label{fig:decoded_vs_gt}
    \vspace{-10pt}
\end{figure*}

\outline{3D data are tr\`es cool}
Three-dimensional (3D) representations of real-life objects are a core tool for vision, robotics, medicine, augmented reality and virtual reality applications.
\outline{Classic encodings are handy but lack semantics}
Recent attempts to encode 3D geometry for use in deep learning include view-based projections, volumetric grids and graphs. 
In this work, we focus on the representation of 3D point clouds. Point clouds are becoming increasingly popular as a homogeneous, expressive and compact representation of surface-based geometry, with the ability to represent geometric details while taking up little space. Point clouds are particularly amenable to simple geometric operations and are a standard 3D acquisition format used by range-scanning devices like LiDARs, the Kinect or iPhone's face ID feature.


All the aforementioned encodings, while effective in their target tasks (e.g. rendering or acquisition), are hard to manipulate directly in their raw form.
For example, na\"ively interpolating between two cars in any of those representations does not yield a representation of an ``intermediate'' car.
Furthermore, these representations are not well suited for the design of generative models via classical statistical methods.
Using them to edit and design new objects involves the construction and manipulation of custom, object-specific parametric models, that link the semantics to the representation.
This process requires significant expertise and effort.


\outline{Deep learning}
Deep learning brings the promise of a {\em data-driven approach}.
In domains where data is plentiful, deep learning tools have eliminated the need for hand-crafting features and models.
Architectures like AutoEncoders (AEs) \cite{rumelhart1988learning,kingma2013auto}
and Generative Adversarial Networks (GANs) \cite{goodfellow2014generative,radford2015unsupervised,che2016mode} are successful at learning data representations and generating realistic samples from complex underlying distributions.
However, an issue with GAN-based generative pipelines is that training them is notoriously hard and unstable~\cite{salimans2016improved}.
In addition, and perhaps more importantly, {\em there is no universally accepted method for the evaluation of generative models}.

\outline{Point clouds new in DL}
In this paper, we explore the use of deep architectures for {\em learning representations} and introduce the first deep {\em generative models} for {\em point clouds}.
Only a handful of deep architectures tailored to 3D point clouds exist in the literature, and their focus is elsewhere: they either aim at classification and segmentation \cite{qi2016pointnet,DBLP:journals/corr/QiYSG17}, or use point clouds {\em only} as an intermediate or output representation \cite{kalogerakis2016_3d-shape-segmentation-with-projective,
fan2016_a-point-set-generation-network-for-3d-object}.
Our specific contributions are:
\vspace{-0.1in}
\begin{itemize}
\item A new AE architecture for point clouds---inspired by recent architectures used for classification \cite{qi2016pointnet}---that can learn compact representations with (i) good reconstruction quality on unseen samples; (ii)  good classification quality via simple methods (SVM), outperforming the state of the art \cite{wu2016_learning-a-probabilistic-latent-space};
(iii) the capacity for meaningful semantic operations, interpolations and shape-completion.
\item The first set of deep generative models for point clouds, able to synthesize point clouds with (i) measurably high fidelity to, and (ii) good coverage of both the training and the held-out data. One workflow that we propose is to first train an AE to learn a latent representation and
then train a generative model in that fixed latent space. The GANs trained in the latent space, dubbed here {\em l-GANs}, are easier to train than raw GANs and achieve superior reconstruction and better coverage of the data distribution.
Multi-class GANs perform almost on par with class-specific GANs when trained in the latent space.
\item A study of various old and new point cloud metrics, in terms of their applicability
(i) as reconstruction objectives for learning good representations;
(ii) for the evaluation of generated samples.
We find that a commonly used metric, Chamfer distance, fails to
identify
certain pathological cases.
\item
Fidelity and coverage metrics for generative models, based on an optimal matching between two different
 collections of point clouds.
Our coverage metric can identify parts of the data distribution that are completely missed by the generative model, something that diversity metrics based on cardinality might fail to capture \cite{arora2017gans}.
\end{itemize}

The rest of this paper is organized as follows: Section~\ref{sec:background}
outlines some background for the basic building blocks of our work.
Section~\ref{sec:evaluations_metrics} introduces our metrics for the evaluation of generative point cloud pipelines.
Section~\ref{sec:models} discusses our architectures for latent representation learning and generation.
In Section~\ref{sec:results}, we perform comprehensive experiments evaluating all of our models both quantitatively and qualitatively. Further results can be found in the Appendix. Last, the code for all our models is publicly available\footnote{\url{http://github.com/optas/latent_3d_points}}.

\section{Background}
\label{sec:background}

In this section we give the necessary background on point clouds, their metrics and the fundamental building blocks that we will use in the rest of the paper.

\subsection{Point clouds}
\paragraph{Definition}
A point cloud represents a geometric shape---typically its  surface---as a set of 3D locations in a Euclidean coordinate frame.
In 3D, these locations are defined by their $x, y, z$ coordinates. Thus, the point cloud representation of an object or scene is a $N\times 3$ matrix, where $N$ is the number of points, referred to as the point cloud resolution.

Point clouds as an input modality present a unique set of challenges when building a network architecture. As an example, the convolution operator---now ubiquitous in image-processing pipelines---requires the input signal to be defined on top of an underlying grid-like structure. Such a structure is not available in raw point clouds, which renders them significantly more difficult to encode than images or voxel grids. Recent classification work on point clouds (PointNet \cite{qi2016pointnet}) bypasses this issue by avoiding convolutions involving groups of points. Another related issue with point clouds as a representation is that they are permutation invariant: any reordering of the rows of the point cloud matrix yields a point cloud that represents the same shape. This property complicates comparisons between two point sets which is needed to define a reconstruction loss. It also creates the need for making the encoded feature permutation invariant.

\paragraph{Metrics}
Two permutation-invariant metrics for comparing unordered point sets have been proposed in the literature~\cite{fan2016_a-point-set-generation-network-for-3d-object}. On the one hand, the  \emph{Earth Mover's} distance (EMD)~\cite{rubner2000_the-earth-movers-distance-as-a-metric} is the solution of a transportation problem which attempts to transform one set to the other. For two equally sized subsets $S_1 \subseteq R^3, S_2 \subseteq R^3$, their EMD is defined by
\begin{displaymath}
	d_{EMD}(S_1,S_2) =  \min\limits_{\phi: S_1 \to S_2} 	\sum\limits_{x\in S_1}\|x - \phi(x)\|_2
\end{displaymath}
where $\phi$ is a bijection.
As a loss, EMD is differentiable almost everywhere.
On the other hand, the \emph{Chamfer} (pseudo)-distance (CD) measures the squared distance between each point in one set to its nearest neighbor in the other set:
\begin{displaymath}
	d_{CH}(S_1,S_2) = \sum\limits_{x\in S_1}\min_{y\in S	_2}\|x-y\|_2^2 + \sum\limits_{y\in S_2}\min_{x\in S_1}\|x - y\|_2^2.
\end{displaymath}
CD is differentiable and compared to EMD more efficient to compute.

\subsection{Fundamental building blocks}

\paragraph{Autoencoders}
One of the main
deep-learning components we use in this paper is the
\emph{AutoEncoder} (AE, inset),
\begin{wrapfigure}{r}{0.50\columnwidth}
  \vspace{-10pt}
    \includegraphics[width=1.0\linewidth]{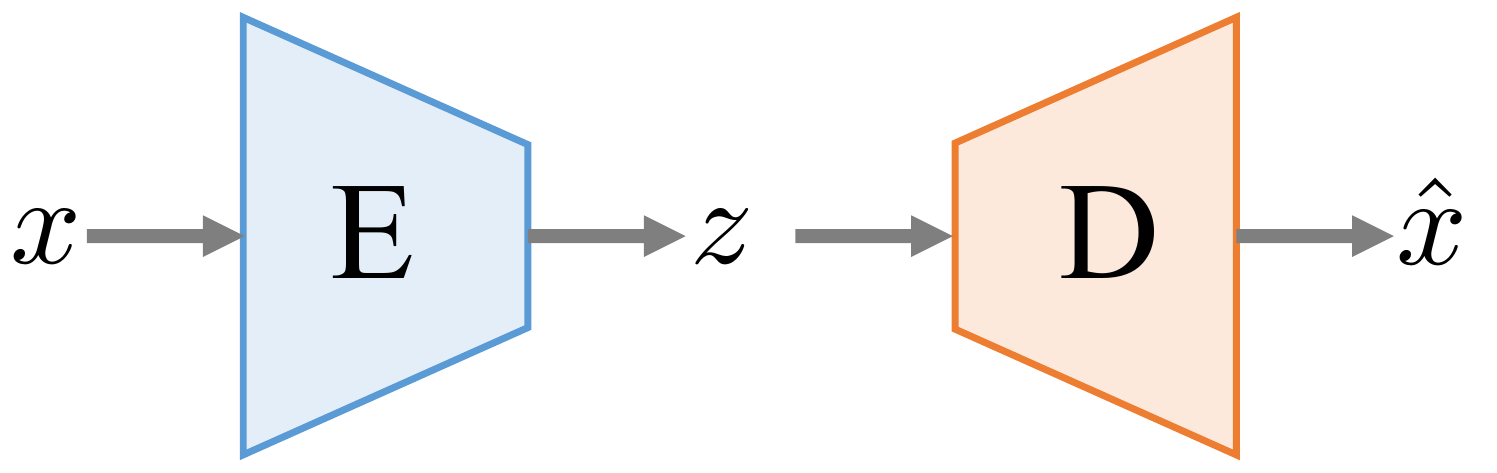}
\vspace{-20pt}
\end{wrapfigure}
which is an architecture that learns
to reproduce its input. AEs can be especially useful, when they contain a narrow {\em bottleneck layer} between  input and output. Upon successful training, this layer provides a low-dimensional representation, or {\em code}, for each data point. The Encoder (E) learns to compress a data point $\vx$ into its latent representation, $\vz$. The Decoder (D) can then produce a reconstruction $\hat{\vx}$, of $\vx$, from its encoded version  $\vz$.

\paragraph{Generative Adversarial Networks}
In this paper we also work with Generative Adversarial Networks (GANs), which are
state-of-the-art generative models.
The basic
architecture (inset) is based on a adversarial game between a {\em generator} (G) and a {\em discriminator} (D).
The generator aims to synthesize samples that
look
\begin{wrapfigure}{r}{0.50\columnwidth}
\vspace{-10pt}
    \includegraphics[width=1.0\linewidth]{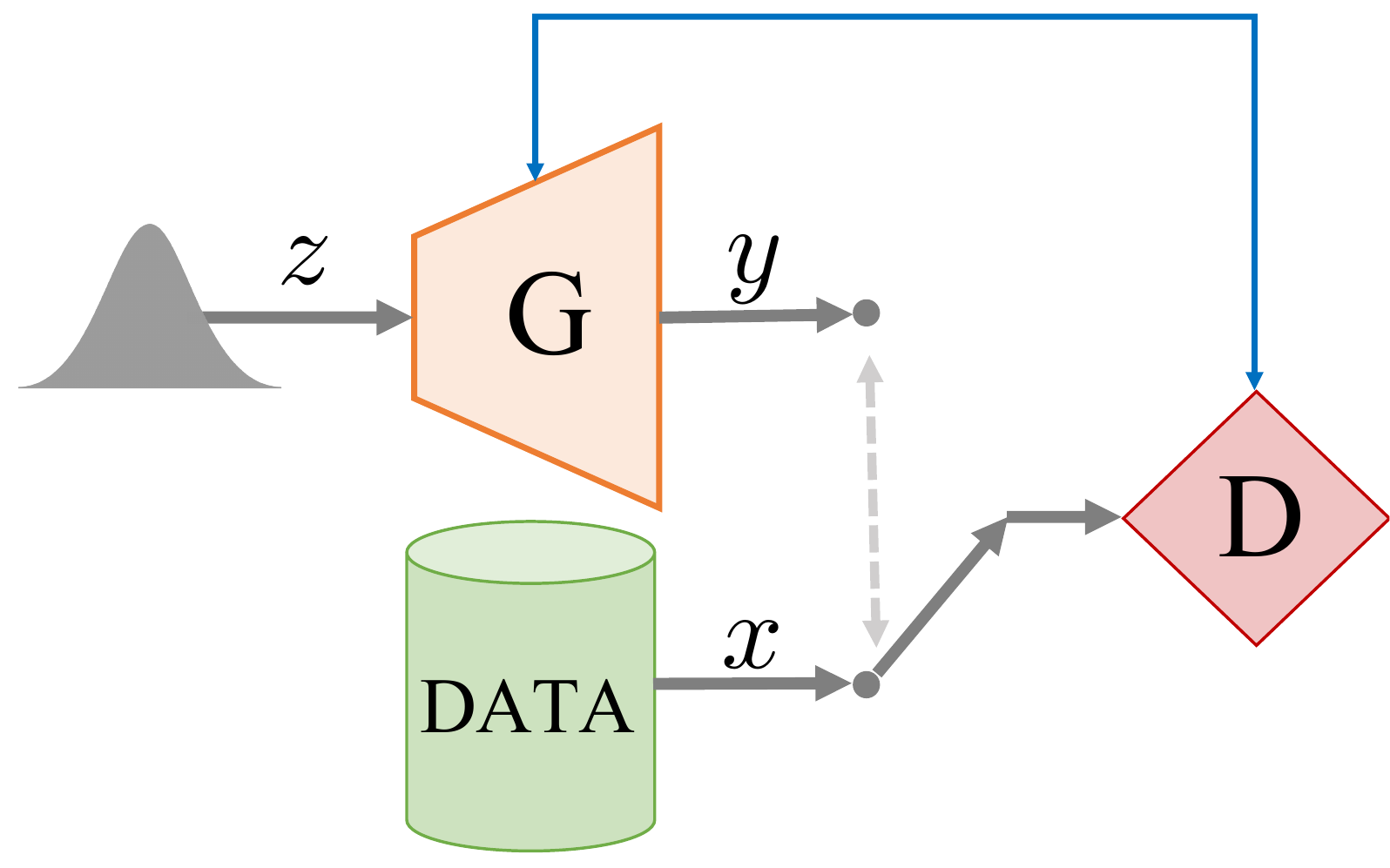}
\vspace{-25pt}
\end{wrapfigure}
indistinguishable from real data
(drawn from $\vx \sim \pdata$)
by passing a randomly drawn sample from a simple distribution $\vz \sim p_z$ through the generator function.
The discriminator is tasked with distinguishing synthesized from real samples.


\paragraph{Gaussian Mixture Model}
A GMM is a probabilistic model for representing a population whose distribution is assumed to be multimodal Gaussian, i.e.\ comprising of multiple subpopulations, where each subpopulation follows a Gaussian distribution. Assuming the number of subpopulations is known, the GMM parameters (means and variances of the Gaussians) can be learned from random samples, using the Expectation-Maximization (EM) algorithm \cite{Dempster:EM}.
Once fitted, the GMM can be used to sample novel synthetic samples.

\section{Evaluation Metrics for Generative Models}
\label{sec:evaluations_metrics}

\begin{figure*}
    \centering
    \includegraphics[height=2.2cm, width=\textwidth]{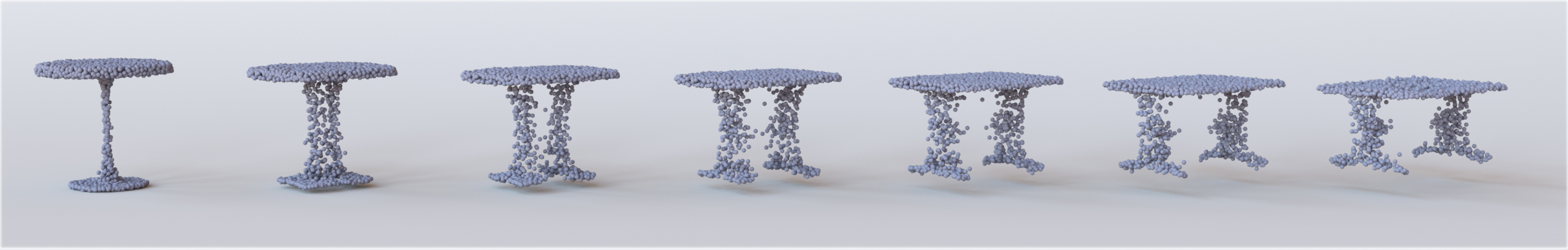}
    \vspace{-20pt}
    \caption{Interpolating between different point clouds, using our latent space representation. More examples for furniture and {\em human-form} objects~\protect{\cite{dfaust:CVPR:2017}} are demonstrated in the Appendix in Figures~\ref{fig:interpolations-2} and~\ref{fig:interpolations-bogo}, respectively.}
    \label{fig:interpolations-1}
    \vspace{-10pt}
\end{figure*}

An important component of this work is the introduction of measures that enable comparisons between two sets of points clouds $A$ and $B$. These metrics are useful for assessing the degree to which point clouds, synthesized or reconstructed, represent the same population as a held-out test set. Our three measures are described below.

\paragraph{JSD}
The Jensen-Shannon Divergence between marginal distributions defined in the Euclidean 3D space. Assuming point cloud data that are axis-aligned and a canonical voxel grid in the ambient space; one can measure the degree to which point clouds of $A$ tend to occupy similar locations as those of $B$. To that end, we count the number of points lying within each voxel across {\em all} point clouds of $A$, and correspondingly for $B$ and report the JSD between the obtained empirical distributions $(P_A, P_B)$:
\begin{equation*}
JSD(P_A \parallel P_B)= \frac{1}{2}D(P_A \parallel M)+\frac{1}{2}D(P_B \parallel M)\nonumber
\end{equation*}
where $M=\frac{1}{2}(P_A+P_B)$ and $D(\cdot \parallel \cdot)$ the KL-divergence between the two distributions \cite{KL-div}.

\paragraph{Coverage} For each point cloud in $A$ we first find its closest neighbor in $B$.
Coverage is measured as the {\em fraction} of the point clouds in $B$ that were matched to point clouds in $A$.
Closeness can be computed using either the CD or EMD point-set distance of Section~\ref{sec:background}, thus yielding two different metrics, COV-CD and COV-EMD. A high coverage score indicates that most of $B$ is roughly represented within $A$.

\paragraph{Minimum Matching Distance (MMD)}
Coverage does not indicate exactly {\em how well} the covered examples (point-clouds) are represented in set $A$; matched examples need not be close.
We need a way to measure the {\em fidelity} of $A$ with respect to $B$. To this end, we match every point cloud of $B$ to the one in $A$ with the minimum distance (MMD) and report the average of distances in the matching.
Either point-set distance can be used, yielding MMD-CD and MMD-EMD. Since MMD relies directly on the distances of the matching, it correlates well with how faithful (with respect to $B$) elements of $A$ are.

\paragraph{Discussion} The complementary nature of MMD and Coverage directly follows from their definitions. The set of point clouds $A$ captures all modes of $B$ with good fidelity when MMD is small {\em and} Coverage is large. JSD is fundamentally different. First, it evaluates the similarity between $A$ and $B$ in coarser way, via marginal statistics. Second and contrary to the other two metrics, it requires pre-aligned data, but is also computationally friendlier. We have found and show experimentally that it correlates well with the MMD, which makes it an efficient alternative for e.g. model-selection, where one needs to perform multiple comparisons between sets of point clouds.

\section{Models for Representation and Generation}
\label{sec:models}

In this section we describe the architectures of our neural networks starting from an autoencoder. Next, we introduce a GAN that works directly with 3D point cloud data, as well as a decoupled approach which first trains an AE and then trains a minimal GAN in the AE's latent space. We conclude with a similar but even simpler solution that relies on classical Gaussian mixtures models.

\subsection{Learning representations of 3D point clouds}
\label{sec:models:ae}
The input to our AE network is a point cloud with 2048 points ($2048\times 3$ matrix), representing a 3D shape. The encoder architecture follows the design principle of ~\cite{qi2016pointnet}: 1-D convolutional layers with kernel size 1 and an increasing number of features; this approach encodes every point {\em independently}. A "symmetric", permutation-invariant function (e.g. a max pool) is placed after the convolutions to produce a joint representation. In our implementation we use 5 1-D convolutional layers, each followed by a ReLU ~\cite{relu} and a batch-normalization layer ~\cite{bnorm}. The output of the last convolutional layer is passed to a feature-wise maximum to produce a $k$-dimensional vector which is the basis for our latent space. Our decoder transforms the latent vector using 3 fully connected layers, the first two having ReLUs, to produce a $2048\times3$ output.
For a permutation invariant objective, we explore both the EMD approximation and the CD (Section~\ref{sec:background}) as our structural losses; this yields two distinct AE models, referred to as AE-EMD and AE-CD. To regularize the AEs we considered various bottleneck sizes, the use of drop-out and on-the-fly augmentations by randomly-rotating the point clouds. The effect of these choices is showcased in the Appendix (Section~\ref{app:ae_details}) along with the detailed training/architecture parameters. In the remainder of the paper, unless otherwise stated, we use an AE with a $128$-dimensional bottleneck layer.

\subsection{Generative models for Point Clouds}
\label{sec:models:generative}

\paragraph{Raw point cloud GAN (r-GAN)}
Our first GAN operates on the raw $2048\times 3$ point set input.
The architecture of the discriminator is identical to the AE (modulo the filter-sizes and number of neurons), without any batch-norm and with leaky ReLUs~\cite{maas2013_rectifier-nonlinearities-improve-neural} instead or ReLUs. The output of the last fully connected layer is fed into a sigmoid neuron. The generator takes as input a Gaussian noise vector and maps it to a $2048\times 3$ output via 5 FC-ReLU layers.

\paragraph{Latent-space GAN (l-GAN)}
For our l-GAN, instead of operating on the raw point cloud input, we pass the data through a pre-trained autoencoder, which is trained separately for each object class with the EMD (or CD) loss function.
Both the generator and the discriminator of the l-GAN then operate on the bottleneck variables of the AE.
Once the training of GAN is over, we convert a code learned by the generator into a point cloud by using the AE's decoder.
Our chosen architecture for the l-GAN, which was used throughout our experiments, is {\em significantly} simpler than the one of the r-GAN. Specifically, an MLP generator of a single hidden layer coupled with an MLP discriminator of two hidden layers suffice to produce measurably good and realistic results.

\paragraph{Gaussian mixture model}
In addition to the l-GANs, we also fit a family of Gaussian Mixture Models (GMMs) on the latent spaces learned by our AEs. We experimented with various numbers of Gaussian components and diagonal or full covariance matrices. The GMMs can be turned into point cloud generators by first sampling the fitted distribution and then using the AE's decoder, similarly to the l-GANs.

\section{Experimental Evaluation}
\label{sec:results}

\begin{figure*}
    \centering
    \includegraphics[height=2.0cm, width=\textwidth]{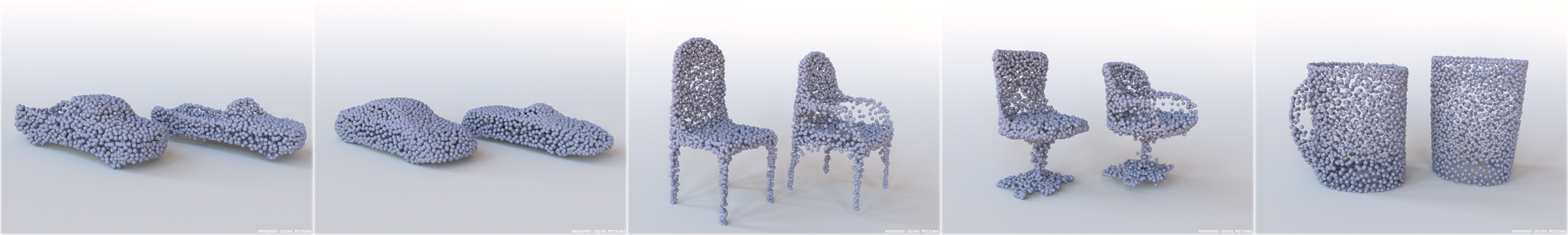}
    \vspace{-20pt}
    \caption{Editing parts in point clouds using simple additive algebra on the AE latent space. Left to right: tuning the appearance of cars towards the shape of convertibles, adding armrests to chairs, removing handle from mug. Note that the height of chairs with armrests is on average 13\% shorter than of chairs without one; which is reflected also in these results.}
    \label{fig:part_editing}
    \vspace{-10pt}
\end{figure*}

In this section we experimentally establish the validity of our proposed evaluation metrics and highlight the merits of the AE-representation (Section~\ref{sec:results:evaluation_ae}) and the generative models (Section~\ref{sec:results:evaluation_generative}).  In all experiments in the main paper, we use shapes from the ShapeNet repository \cite{chang2015_shapenet:-an-information-rich-3d-model-repository}, that are axis aligned and centered into the unit sphere. To convert these shapes (meshes) to point clouds we uniformly sample their faces in proportion to their area. Unless otherwise stated, we train models with point clouds from a single object class and work with train/validation/test sets of an 85\%-5\%-10\% split. When reporting JSD measurements we use a $28^3$ regular voxel grid to compute the statistics. 

\subsection{Representational power of the AE}
\label{sec:results:evaluation_ae}
We begin with  demonstrating the merits of the proposed AE.
First we report
its generalization ability as measured using the MMD-CD and MMD-EMD metrics. Next, we utilize its latent codes to do semantically meaningful operations. Finally, we use the latent representation to train SVM classifiers and report the attained classification scores.

\paragraph{Generalization ability.} \label{ae_ge}
Our AEs are able to reconstruct unseen shapes with quality almost as good as that of the shapes that were used for training. In Fig.~\ref{fig:decoded_vs_gt} we use our AEs to encode unseen samples from the \emph{test} split (the left of each pair of images) and then decode them and compare them visually to the input (the right image). To support our visuals quantitatively, in Table~\ref{table:aes_on_generative_measurements} we report the MMD-CD and MMD-EMD between reconstructed point clouds and their corresponding ground-truth in the train and test datasets of the chair object class. The generalization gap under our metrics is small; to give a sense of scale for our reported numbers, note that the MMD is $0.0003$ and $0.033$ under the CD and EMD respectively between two versions of the test set that only differ by the randomness introduced in the point cloud sampling. Similar conclusions regarding the generalization ability of the AE can be made based on the reconstruction loss attained for each dataset (train or test) which is shown in Fig.~\ref{fig:optimal_bottleneck} of the Appendix.

\begin{table}[htb]
  \small
  \centering
    \begin{tabularx}{\columnwidth}{C{.6} C{1.} C{1.} C{1.2} C{1.2}}
        \hline
        \multirow{2}{*}{AE} & \multicolumn{2}{c}{MMD-CD} & \multicolumn{2}{c}{MMD-EMD}\\
        \cmidrule(lr){2-3} \cmidrule(l){4-5}
        loss &Train & Test & Train & Test\\
        \hline
        \hline
        CD  & {\bf 0.0004} & {\bf 0.0012} & 0.068 & 0.075\\
        EMD & 0.0005 & 0.0013 & {\bf 0.042} & {\bf 0.052}\\
        \hline
    \end{tabularx}
    \vspace{-5pt}
    \caption{Generalization of AEs as captured by MMD. Measurements for reconstructions on the training and test splits for an AE trained with either the CD or EMD loss and data of the chair class; Note how the MMD favors the AE that was trained with the same loss as the one used by the MMD to make the matching.}
    \label{table:aes_on_generative_measurements}
\end{table}

\paragraph{Latent space and linearity.} \label{ae_linear}
Another argument against under/over-fitting can be made by showing that the learned representation is amenable to intuitive and semantically rich operations. As it is shown in several recent works, well trained neural-nets learn a latent representation where additive linear algebra works to that purpose \cite{word2vec, shape2vec}. First, in Fig.~\ref{fig:interpolations-1} we show linear interpolations, in the latent space, between the left and right-most geometries. Similarly, in Fig.~\ref{fig:part_editing} we alter the input geometry (left) by adding, in latent space, the mean vector of geometries with a certain characteristic (e.g., convertible cars or cups without handles). Additional operations (e.g. shape analogies) are also possible, but due to space limitations we illustrate and provide the details in the Appendix (Section~\ref{app:applications_ae}) instead. These results attest to the smoothness of the learned space but also highlight the intrinsic capacity of point clouds to be smoothly morphed.

\begin{figure*}[t]
    \centering
    \includegraphics[width=\textwidth]{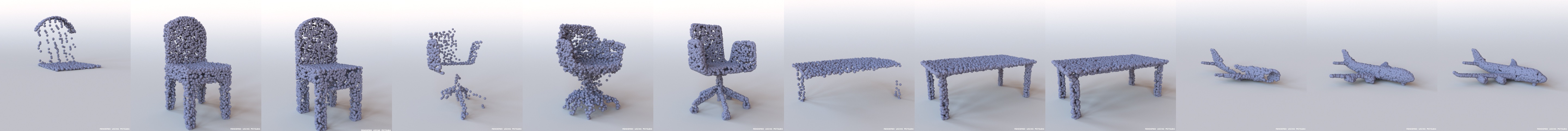}
    \vspace{-20pt}
    \caption{Point cloud {\em completions} of a network trained with partial and complete (input/output) point clouds and the EMD loss. Each triplet shows the partial input from the test split (left-most), followed by the network's output (middle) and the complete ground-truth (right-most).}
    \label{fig:completions-main}
    \vspace{-1pt}
\end{figure*}

\paragraph{Shape completions.} \label{ae_completions}
Our proposed AE architecture can be used to tackle the problem of shape completion with minimal adaptation. Concretely, instead of feeding and reconstructing the same point cloud, we can feed the network with an {\em incomplete} version of its expected output. Given proper training data, our network learns to complete severely partial point clouds. Due to space limitations we give the exact details of our approach in the Appendix (Section~\ref{app:shape_completions}) and demonstrate some achieved completions in Fig.~\ref{fig:completions-main} of the main paper.

\paragraph{Classification.} \label{ae_svm}
Our final evaluation for the AE's design and efficacy is done by using the learned latent codes as features for classification. For this experiment to be meaningful, we train an AE across all different shape categories: using 57,000 models from 55 categories of man-made objects. Exclusively for this experiment, we use a bottleneck of $512$ dimensions and apply random rotations to the input point clouds along the gravity axis. To obtain features for an input 3D shape, we feed its point cloud into the AE and extract the bottleneck activation vector. This vector is then classified by a linear SVM trained on the de-facto 3D classification benchmark of ModelNet~\cite{wu2015_3d-shapenets:-a-deep-representation-for-volumetric}. Table~\ref{table:unsup_classification} shows comparative results. Remarkably, in the ModelNet10 dataset, which includes classes (chairs, beds etc.) that are populous in ShapeNet, our simple AE significantly outperforms the state of the art \cite{wu2016_learning-a-probabilistic-latent-space} which instead uses several layers of a GAN to derive a $7168$-long feature. In Fig.~\ref{fig:confusion_matrix} of the Appendix we include the confusion matrix of the classifier evaluated on our latent codes on ModelNet40 --  the confusion happens between particularly similar geometries: a dresser vs. a nightstand or a flower-pot vs. a plant. The nuanced details that distinguish these objects may be hard to learn without stronger supervision.


\begin{table}[htb]
    \small
    \centering
    \vspace{5pt}
    \begin{tabularx}{\columnwidth}{C{1} C{1} C{1} C{1} C{1} C{1} C{1} C{1}}
        \hline
         & A & B & C & D & E & {ours EMD} & {ours CD}\\
        \hline
        \hline
        MN10 & 79.8 & 79.9 &  -     & 80.5 & 91.0 & {\bf95.4} & \bf{95.4}\\
        MN40 & 68.2 & 75.5 & 74.4 & 75.5 & 83.3 & 84.0      & {\bf84.5}\\
        \hline
  \end{tabularx}
    \vspace{-5pt}
    \caption{Classification performance (in \%) on ModelNet10/40. Comparing to A:~SPH \cite{Kazhdan_03_SGP}, B:~LFD \cite{Chen_03_CGF}, C:~T-L-Net \cite{girdhar1eccv}, D:~VConv-DAE \cite{sharma16eccvw}, E:~3D-GAN \cite{wu2016_learning-a-probabilistic-latent-space}.}
    \label{table:unsup_classification}
    \vspace{-5pt}
\end{table}

\begin{figure*}[ht]
    \centering
    \includegraphics[width=\textwidth]{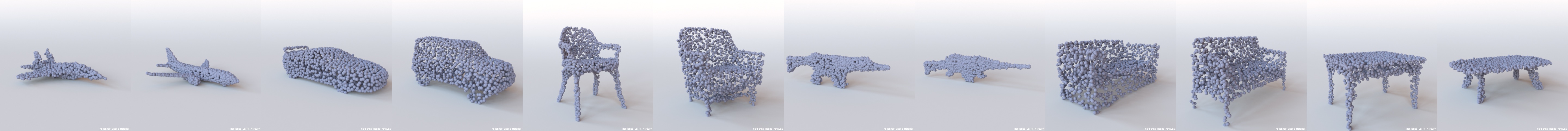}
    \includegraphics[width=\textwidth]{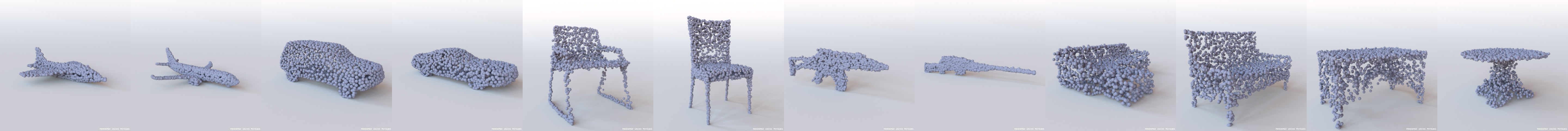}
    \vspace{-20pt}
    \caption{Synthetic point clouds generated by samples produced with l-GAN (top) and 32-component GMM (bottom), both trained on the latent space of an AE using the EMD loss.}
    \label{fig:generative_l_w_gan_and_gmm}
\end{figure*}

\subsection{Evaluating the generative models}

\label{sec:results:evaluation_generative}
Having established the quality of our AE, we now demonstrate the merits and shortcomings of our generative pipelines and establish one more successful application for the AE's learned representation.
First, we conduct a comparison between our generative models followed by a comparison between our latent GMM generator and the state-of-the-art 3D voxel generator. Next, we describe how Chamfer distance can yield misleading results in certain pathological cases that r-GANs tends to produce. Finally, we show the benefit of working with a pre-trained latent representation in multi-class generators.

\begin{figure}
    \centering
    \vspace{10pt}
    \includegraphics[width=.5\columnwidth]{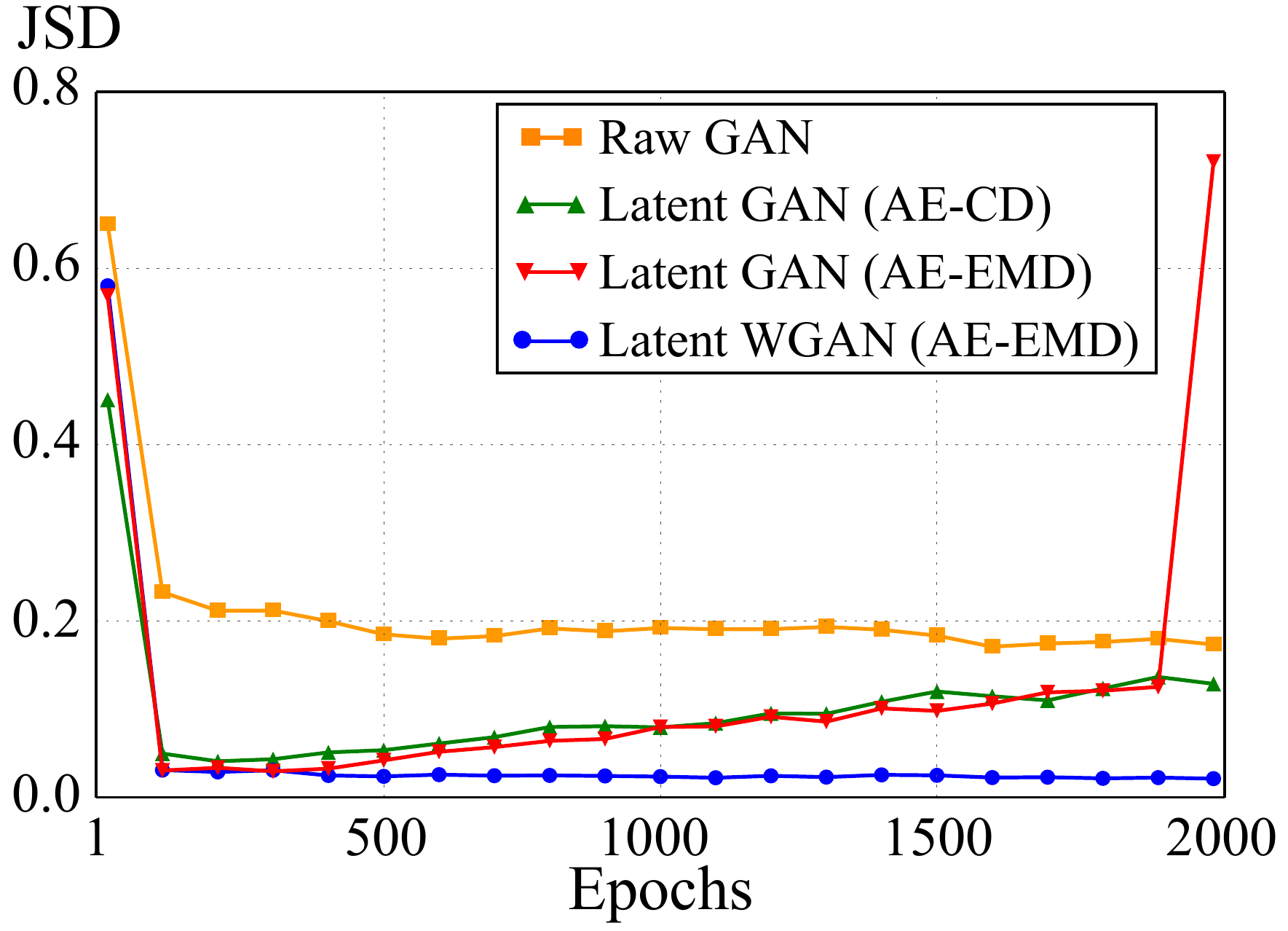}
    \hspace{-5pt}
    \includegraphics[width=.5\columnwidth]{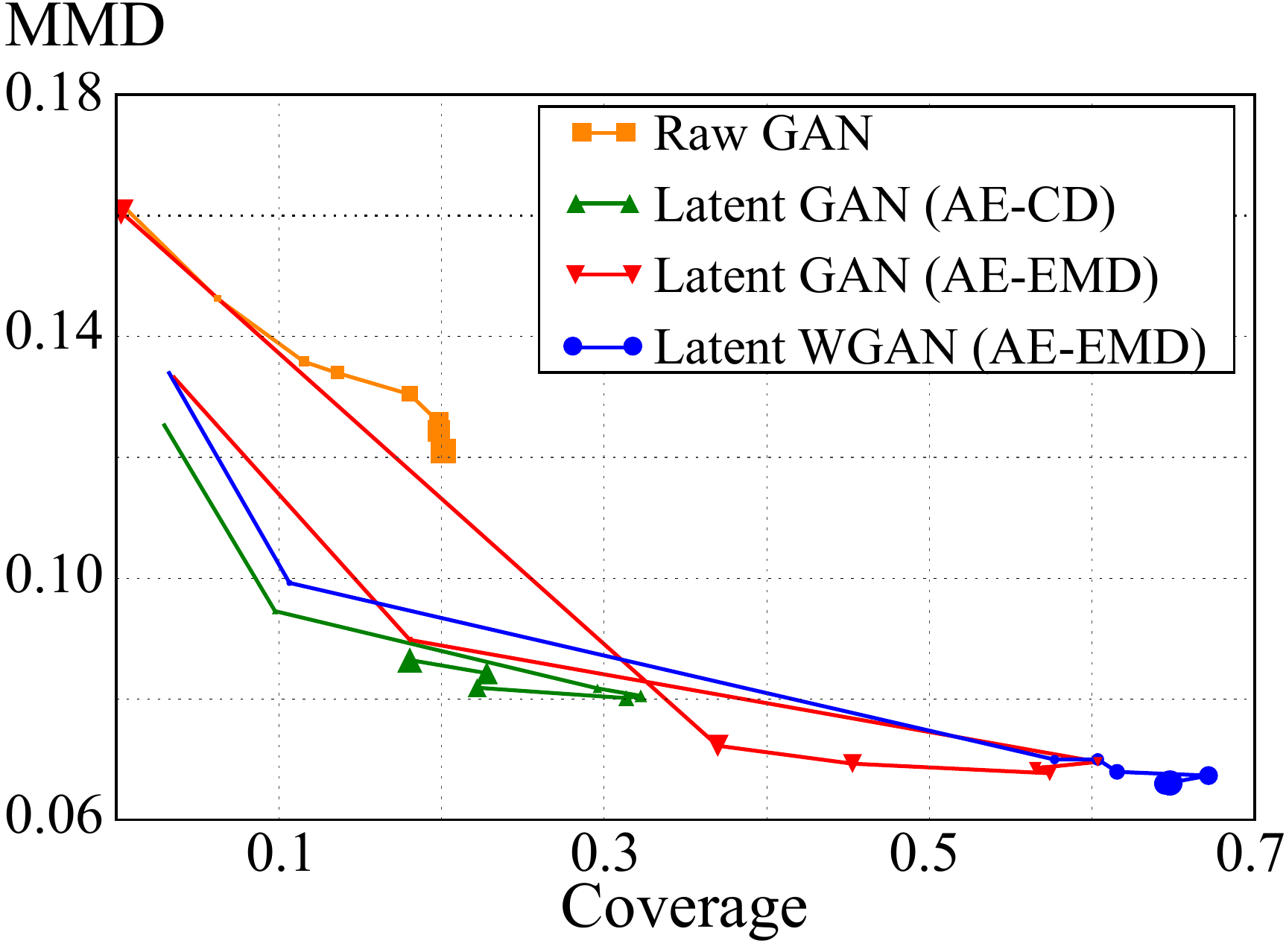}
    \vspace{-12pt}    
    \caption{Learning behavior of the GANs, in terms of coverage / fidelity to the ground truth  \textbf{test} dataset. Left -- the JSD distance between the ground truth test set and synthetic datasets generated by the GANs at various epochs of training. Right -- EMD based MMD/Coverage:  curve markers indicate epochs 1, 10, 100, 200, 400, 1000, 1500, 2000, with {\em larger symbols denoting later epochs.}}
    \label{fig:training_trends}
\end{figure}

\paragraph{Comparison of our different generative models}
For this study, we train five generators with point clouds of the \emph{chair} category. First, we establish two AEs trained with the CD or EMD loss respectively---referred to as AE-CD and AE-EMD and train an l-GAN in each latent space with the non-saturating loss of \citet{goodfellow2014generative}. In the space learned by the AE-EMD we train two additional models: an identical (architecture-wise) l-GAN that utilizes the Wasserstein objective with gradient-penalty \cite{GulrajaniAADC17} and a family of GMMs with a different number of means and structures of covariances. We also train an r-GAN directly on the point cloud data.

Fig.~\ref{fig:training_trends} shows the JSD (left) and the MMD and Coverage (right) between the produced synthetic datasets and the held-out {\em test} data for the GAN-based models, as training proceeds. Note that the r-GAN struggles to provide good coverage and good fidelity of the test set; which alludes to the well-established fact that end-to-end GANs are generally difficult to train. The l-GAN (AE-CD) performs better in terms of fidelity with much less training, but its coverage remains low. Switching to an EMD-based AE for the representation and otherwise using the same latent GAN architecture (l-GAN, AE-EMD), yields a dramatic improvement in coverage and fidelity. Both l-GANs though suffer from the known issue of mode collapse: half-way through training, first coverage starts dropping with fidelity still at good levels, which implies that they are overfitting a small subset of the data. Later on, this is followed by a more catastrophic collapse, with coverage dropping as low as 0.5\%. Switching to a latent WGAN largely eliminates this collapse, as expected.

In Table~\ref{table:chair_test_data_response}, we report measurements for all generators based on the epoch (or underlying GMM parameters) that has minimal JSD between the generated samples and the validation set. To reduce the sampling bias of these measurements each generator produces a set of synthetic samples that is $3\times$ the population of the comparative set (test or validation) and repeat the process $3$ times and report the averages. The GMM selected by this process has $32$ Gaussians and a full covariance. As shown in Fig.~\ref{fig:jsd_of_gmms_and_cov} of the Appendix, GMMs with full covariances perform much better than those that have diagonal structure and \tilde 20 Gaussians suffice for good results. Last, the first row of Table~\ref{table:chair_test_data_response} shows  a baseline model that memorizes a random subset of the training data of the same size as the other generated sets.

{\em Discussion.} The results of Table~\ref{table:chair_test_data_response} agree with the trends shown in Fig.~\ref{fig:training_trends} and further verify the superiority of the latent-based approaches and the relative gains of using an AE-EMD vs. an AE-CD. Moreover they demonstrate that a simple GMM can achieve results of comparable quality to a latent WGAN. Lastly, it is worth noting how the GMM has achieved similar fidelity as that of the perfect/memorized chairs and with almost as good coverage. Table 8 of the supplementary shows the same performance-based conclusions when our metrics are evaluated on the {\em train} split.

\begin{table}[htb]
  \small
  \centering
    \begin{tabularx}{\columnwidth}{C{1} C{1} C{1} C{1} C{1} C{1} C{1}}    
        \hline
        Model & Type & JSD & MMD-CD & MMD-EMD  & COV-EMD & COV-CD\\
        \hline
        \hline
        \small A  & MEM & 0.017  &0.0018 & 0.063  &78.6 & 79.4\\
        \hline
        \small B  & RAW  & 0.176  &0.0020 & 0.123  &19.0  &52.3 \\
        \small C  & CD   & 0.048  &0.0020 & 0.079  &32.2  &59.4 \\
        \small D  & EMD  & 0.030  &0.0023 & 0.069  &57.1  &59.3 \\
        \small E  & EMD  & 0.022  &0.0019 & 0.066  &66.9  &67.6 \\
        \small F  & GMM  & {\bf 0.020}  &{\bf 0.0018} & {\bf 0.065}  & {\bf 67.4}  & {\bf 68.9}\\
        \hline
  \end{tabularx}
    \caption{Evaluating 5 generators on the {\em test} split of the chair dataset on epochs/models selected via minimal JSD on the validation-split. We report: A:~sampling-based memorization baseline, B:~r-GAN, C:~l-GAN (AE-CD), D:~l-GAN (AE-EMD) , E:~l-WGAN (AE-EMD), F:~GMM (AE-EMD).}
    \label{table:chair_test_data_response}
\end{table}

\begin{figure}
    \centering
    \includegraphics[width=\linewidth]{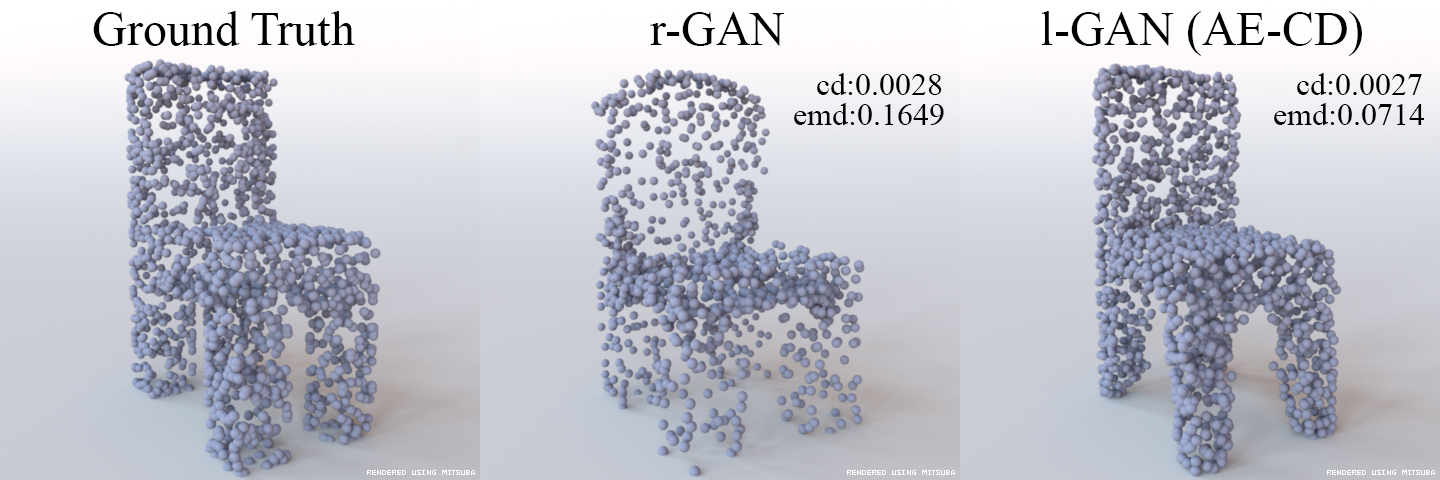}
    \includegraphics[width=\linewidth]{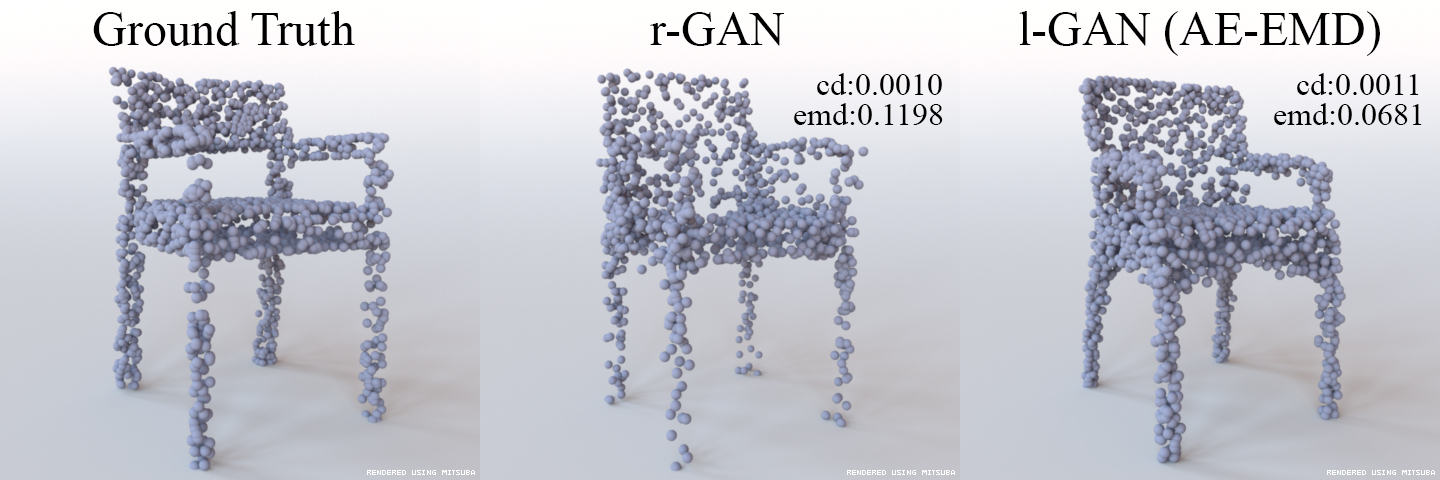}
    \vspace{-20pt}
    \caption{The CD distance is less faithful than EMD to visual quality of synthetic results; here, it favors r-GAN results, due to the overly high density of points in the seat part of the synthesized point sets.
    }
        \label{fig:hotspots_emd_chamfer}
\end{figure}

\paragraph{Chamfer's blindness, r-GAN's hedging.}
An interesting observation regarding r-GAN can be made in Table~\ref{table:chair_test_data_response}. The JSD and the EMD based metrics strongly favor the latent-approaches, while the Chamfer-based ones are much less discriminative.
To decipher this discrepancy we did an extensive qualitative inspection of the r-GAN samples and found many cases of point clouds that were over-populated in locations, that on average, {\em most} chairs have mass. This hedging of the r-GAN is particularly hard for Chamfer to penalize since one of its two summands can become significantly small and the other can be only moderately big by the  presence of a few sparsely placed points in the non-populated locations. Figure~\ref{fig:hotspots_emd_chamfer} highlights this point. For a ground-truth point cloud we retrieve its nearest neighbor, under the CD, in synthetically generated sets produced by the r-GAN and the l-GAN and in-image numbers report their CD and EMD distances from it. Notice how the CD fails to distinguish the inferiority of the r-GAN samples while the EMD establishes it. This blindness of the CD metric to only partially good matches, has the additional side-effect that the CD-based coverage is consistently bigger than the EMD-based one.

\begin{figure*}
    \centering
    \includegraphics[width=\textwidth]{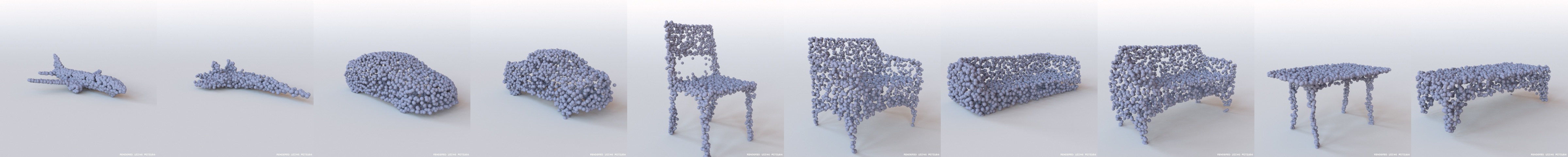}
    \vspace{-20pt}
    \caption{Synthetic point clouds produced with l-WGANs trained in the latent space of an AE-EMD trained on a {\em multi-class} dataset.}
      \label{fig:multi_class}
    \vspace{-10pt}
\end{figure*}

\paragraph{Comparisons to voxel generators.}
\begin{table}[ht]
    \begin{center}
    \begin{tabularx}{\columnwidth}{c *{4}{Y}}
        \hline
        \multirow{2}{*}{Class} & \multicolumn{2}{c}{Fidelity} & \multicolumn{2}{c}{Coverage}\\
        \cmidrule(lr){2-3} \cmidrule(l){4-5}
        & A& Ours &A& Ours\\
        \hline
        \hline
        \emph{car}       &  0.059 & {\bf 0.041} & 28.6 & {\bf 65.3} \\
        \emph{rifle}     &  0.051 & {\bf 0.045} & 69.0 & {\bf 74.8} \\
        \emph{sofa}      &  0.077 & {\bf 0.055} & 52.5 & {\bf 66.6} \\
        \emph{table}     &  0.103 & {\bf 0.061} & 18.3 & {\bf 71.1} \\
        \hline
    \end{tabularx}
    \vspace{-5pt}
    \caption{Fidelity (MMD-EMD) and coverage (COV-EMD) comparison between A:~\citet{wu2016_learning-a-probabilistic-latent-space} and our GMM generative model on the {\em test} split of each class. Note that Wu et al. uses {\em all} models of each class for training contrary to our generators.}
    \label{table:emd_based_wu_comparison}
    \end{center}
\end{table}

Generative models for other 3D modalities, like voxels, have been recently proposed \cite{wu2016_learning-a-probabilistic-latent-space}.
One interesting question is: if point clouds are our target modality, does it make sense to use voxel generators and then convert to point clouds?
This experiment  answers this question in the negative.
First, we make a comparison using a latent GMM which is trained in conjunction with an AE-EMD. Secondly, we build an AE which operates with {\em voxels} and fit a GMM in the corresponding latent space. In both cases, we use 32 Gaussians and a full covariance matrix for these GMMs. To use our point-based metrics, we convert the output of \cite{wu2016_learning-a-probabilistic-latent-space} and our voxel-based GMM into meshes which we sample to generate point clouds. To do this conversion we use the marching-cubes \cite{Lewiner03efficientimplementation} algorithm with an isovalue of $0.1$ for the former method (per authors' suggestions) and $0.5$ for our voxel-AE. We also constrain each mesh to be a single connected component as the vast majority of ground-truth data are.


Table~\ref{table:emd_based_wu_comparison} reveals how our point-based GMM trained with a class specific AE-EMD fares against \cite{wu2016_learning-a-probabilistic-latent-space} on four object classes for which the authors have made their (also class-specific) models publicly\footnote{\url{http://github.com/zck119/3dgan-release}} available. Our approach is consistently better, with a coverage boost that can be as large as $4\times$ and an almost $2\times$ improved fidelity (case of table). This is despite the fact that \cite{wu2016_learning-a-probabilistic-latent-space} uses {\emph all} models of each class for training, contrary to our generators that never had access to the underlying test split.

Table~\ref{table:latent_voxel_generators} reveals the performance achieved by pre-training a {\em voxel}-based AE for the chair class. Observe how by working with a voxel-based latent space, aside of making comparisons more direct to \cite{wu2016_learning-a-probabilistic-latent-space} (e.g. we both convert output voxels to meshes), we also establish significant gains in terms of coverage and fidelity.


\begin{table}[htb]
    \small
    \centering
    \begin{tabularx}{\columnwidth}{C{.2} C{1.2} C{1.4} C{1.} C{1.2}}
        \hline
        {}    & MMD-CD     & MMD-EMD   & COV-CD & COV-EMD \\
        \hline
        \hline
        A    & 0.0046      & 0.091       & 19.6      & 22.4\\
        Ours & \bf{0.0025} & \bf{0.072}  & \bf{60.3} & \bf{64.8} \\
        \hline
    \end{tabularx}
    \vspace{-5pt}
    \caption{MMD and Coverage metrics evaluated on the output of voxel-based methods at resolution $64^3$, matched against the chair {\em test} set, using the same protocol as in Table\ref{table:chair_test_data_response}. Comparing: A:~``raw'' $64^3$-voxel GAN \cite{wu2016_learning-a-probabilistic-latent-space} and a ~latent $64^3$-voxel GMM.}
    \label{table:latent_voxel_generators}
\end{table}

\paragraph{Qualitative results}
In Fig.~\ref{fig:generative_l_w_gan_and_gmm}, we show some synthetic results produced by our l-GANs and the 32-component GMM. We notice high quality results from either model. The shapes corresponding to the 32 means of the Gaussian components can be found in the Appendix (Fig.~\ref{fig:gmm_means_reduced_quality}), as well as results using the r-GAN (Fig.\ref{fig:generative_r_gan}). 

\paragraph{Multi-class generators} Finally, we compare between class specific and class agnostic generators. In Table~\ref{table:single_vs_multi_class} we report the MMD-CD for l-WGANs trained in the space of either a dedicated (per-class) AE-EMD or with an AE-EMD trained with all listed object classes. It turns out that the l-WGANs produce perform similar results in either space. Qualitative comparison (Fig.~\ref{fig:multi_class}) also reveals that by using a multi-class AE-EMD we do not sacrifice much in terms of visual quality compared to the dedicated AEs.

\begin{table}[htb]
  \small
  \centering
  \vspace{6pt}
    \begin{tabularx}{\columnwidth}{C{.2} C{1.2} C{.9} C{1} C{1} C{1.2}  C{1.3}  C{1.2} }
        \hline
        & \emph{airplane} & \emph{car} & \emph{chair} & \emph{sofa} & \emph{table} & average & multi-class\\
        \hline
        \hline
        Tr & 0.0004 & 0.0006 & 0.0015 & 0.0011 & 0.0013 & {\bf 0.0010}  & 0.0011 \\
        Te  & 0.0006 & 0.0007 & 0.0019 & 0.0014 & 0.0017 & {\bf 0.0013} & 0.0014 \\
        \hline
  \end{tabularx}
  \vspace{-5pt}
  \caption{MMD-CD measurements for l-WGANs trained on the latent spaces of dedicated (left 5 columns) and multi-class  EMD-AEs (right column). Also shown is  the weighted average of the per-class values, using the number of train (Tr) resp. test (Te) examples of each class as weights. All l-WGANs use the model parameter resulted by 2000 epochs of training.}
  \label{table:single_vs_multi_class}
\end{table}


\section{Related Work}
Recently, deep learning architectures for view-based projections \cite{su2015_multi-view-convolutional-neural-networks,wei2016_dense-human-body-correspondences,kalogerakis2016_3d-shape-segmentation-with-projective}, volumetric grids \cite{qi2016_volumetric-and-multi-view-cnns-for-object,wu2015_3d-shapenets:-a-deep-representation-for-volumetric,hegde2016fusionnet} and graphs \cite{bruna2013_spectral-networks-and-locally-connected,henaff2015_deep-convolutional-networks-on-graph-structured,defferrard2016_convolutional-neural-networks-on-graphs,yi2016_syncspeccnn:-synchronized-spectral-cnn-for-3d-shape}
have  appeared in the 3D machine learning literature.

A few recent works (\cite{wu2016_learning-a-probabilistic-latent-space}, \cite{wang2016efficient}, \cite{girdhar1eccv}, \cite{brock2016_generative-and-discriminative-voxel-modeling}, \cite{maimaitimin2017_stacked-convolutional-auto-encoders-for-surface}, \cite{zhu2016_deep-learning-representation-using}) have explored generative and discriminative representations for geometry.
They operate on different modalities, typically voxel grids or view-based image projections. To the best of our knowledge, our work is the first to study such representations for point clouds.

Training Gaussian mixture models (GMM) in the latent space of an autoencoder is closely related to VAEs \cite{kingma2013auto}. One documented issue with VAEs is over-regularization:
the regularization term associated with the prior, is often so strong that reconstruction quality suffers \cite{bowman2015generating,sonderby2016train,kingma2016improving,dilokthanakul2016deep}.
The literature contains methods that start only with a reconstruction penalty and slowly increase the weight of the regularizer. An alternative approach is based on adversarial autoencoders \cite{makhzani2015adversarial} which use a GAN to implicitly regularize the latent space of an AE.

\section{Conclusion}
We presented a novel set of architectures for 3D point cloud representation learning and generation. Our results show good generalization to unseen data and our representations encode meaningful semantics. In particular our generative models are able to produce faithful samples and cover most of the ground truth distribution. Interestingly, our extensive experiments
show that the best generative model for point clouds is a GMM trained in the fixed latent space of an AE. While this might not be a universal result, it suggests that simple classic tools should not be dismissed. A thorough investigation on the conditions under which simple latent GMMs are as powerful as adversarially trained models would be of significant interest.

\newpage
\section*{Acknowledgements}
The authors wish to thank all the anonymous reviewers for their insightful comments and suggestions. Lin Shao and Matthias Nie{\ss}ner for their help with the shape-completions and Fei Xia for his suggestions on the evaluation metrics. Last but not least, they wish to acknowledge the support of NSF grants NSF IIS-1528025 and DMS-1546206, ONR MURI grant N00014-13-1-0341, a Google Focused Research award, and a gift from Amazon Web Services for Machine Learning Research. 

\bibliography{00-main.bbl}
\bibliographystyle{icml2018}

\newpage
\appendix
\section {AE Details}
\label{app:ae_details}

The encoding layers of our AEs were implemented as 1D-convolutions with ReLUs, with kernel size of $1$ and stride of $1$, i.e. treating each 3D point independently. Their decoding layers, were MLPs built with FC-ReLUs. We used Adam~\cite{Adam} with initial learning rate of $0.0005$, $\beta_1$ of $0.9$ and a batch size of $50$ to train all AEs.

\subsection{AE used for SVM-based experiments}
For the AE mentioned in the SVM-related experiments of Section~5.1 of the main paper, we used an encoder with $128, 128, 256$ and $512$ filters in each of its layers and a decoder with $1024, 2048, 2048\times3$ neurons, respectively. Batch normalization was used between every layer. We also used online data augmentation by applying random rotations along the gravity-(z)-axis to the input point clouds of each batch. We trained this AE for $1000$ epochs with the CD loss and for $1100$ with the EMD.

\subsection{All other AEs}
For all other AEs, the encoder had $64, 128, 128, 256$ and $k$ filters at each layer, with $k$ being the bottle-neck size. The decoder was comprised by $3$ FC-ReLU layers with $256, 256, 2048\times3$ neurons each. We trained these AEs for a maximum of $500$ epochs when using single class data and $1000$ epochs for the experiment involving $5$ shape classes (end of Section~5.2, main paper).

\subsection{AE regularization}

\begin{figure}
    \centering
    \includegraphics[width=.49\columnwidth]{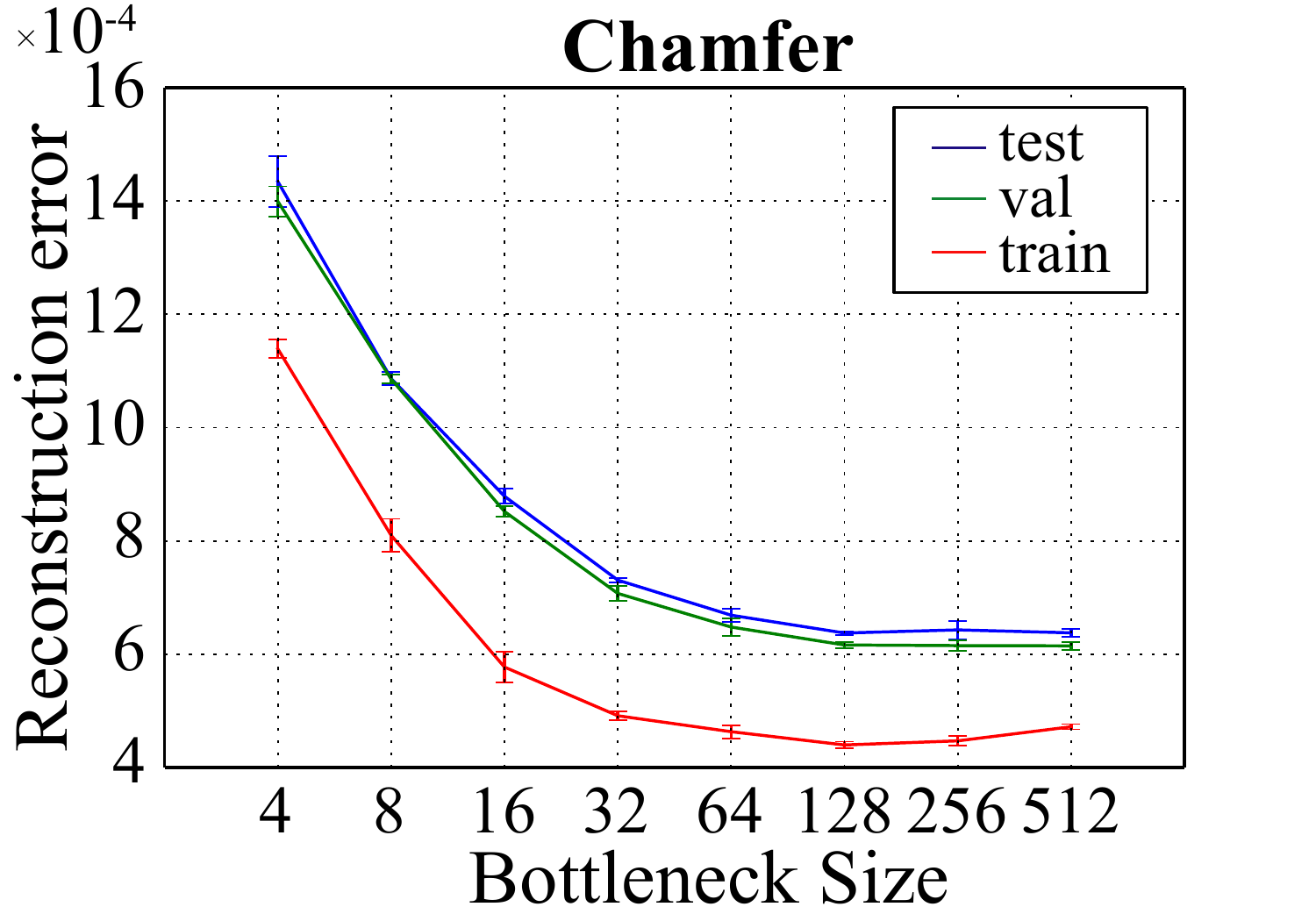}
    \includegraphics[width=.49\columnwidth]{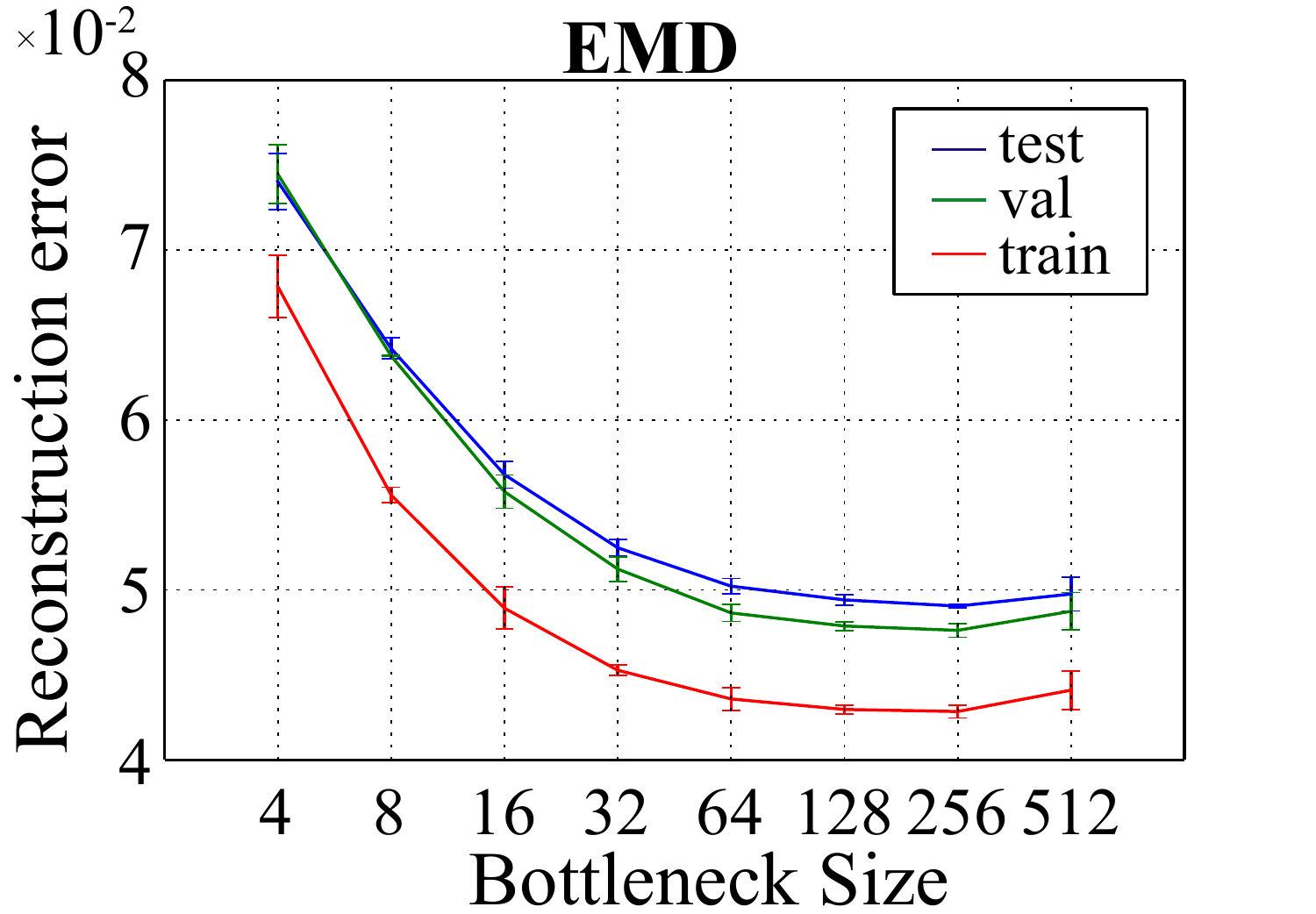}
    \caption{The bottleneck size was fixed at 128 in all single-class experiments by observing the reconstruction loss of the AEs, shown here for various bottleneck sizes, when training with the data of the chair class.}
    \label{fig:optimal_bottleneck}
\end{figure}

To determine an appropriate size for the latent-space, we constructed 8 (otherwise architecturally identical) AEs with bottleneck sizes $k \in \{4, 8\dots, 512\}$ and trained them with point clouds of the chair object class, under the two losses (Fig.~\ref{fig:optimal_bottleneck}). We repeated this procedure with pseudo-random weight initializations three times  and found that $k=128$ had the best generalization error on the test data, while achieving minimal reconstruction error on the train split.

{\bf Remark.} Different AE setups brought no noticeable advantage over our main architecture. Concretely, adding drop-out layers resulted in worse reconstructions and using batch-norm on the encoder {\em only}, sped up training and gave us slightly better generalization error when the AE was trained with single-class data. Exclusively, for the SVM experiment of Section 5.1 of the main paper we randomly rotate the input chairs to promote latent features that are rotation-invariant.

\begin{figure}
    \centering
    \includegraphics[width=\columnwidth]{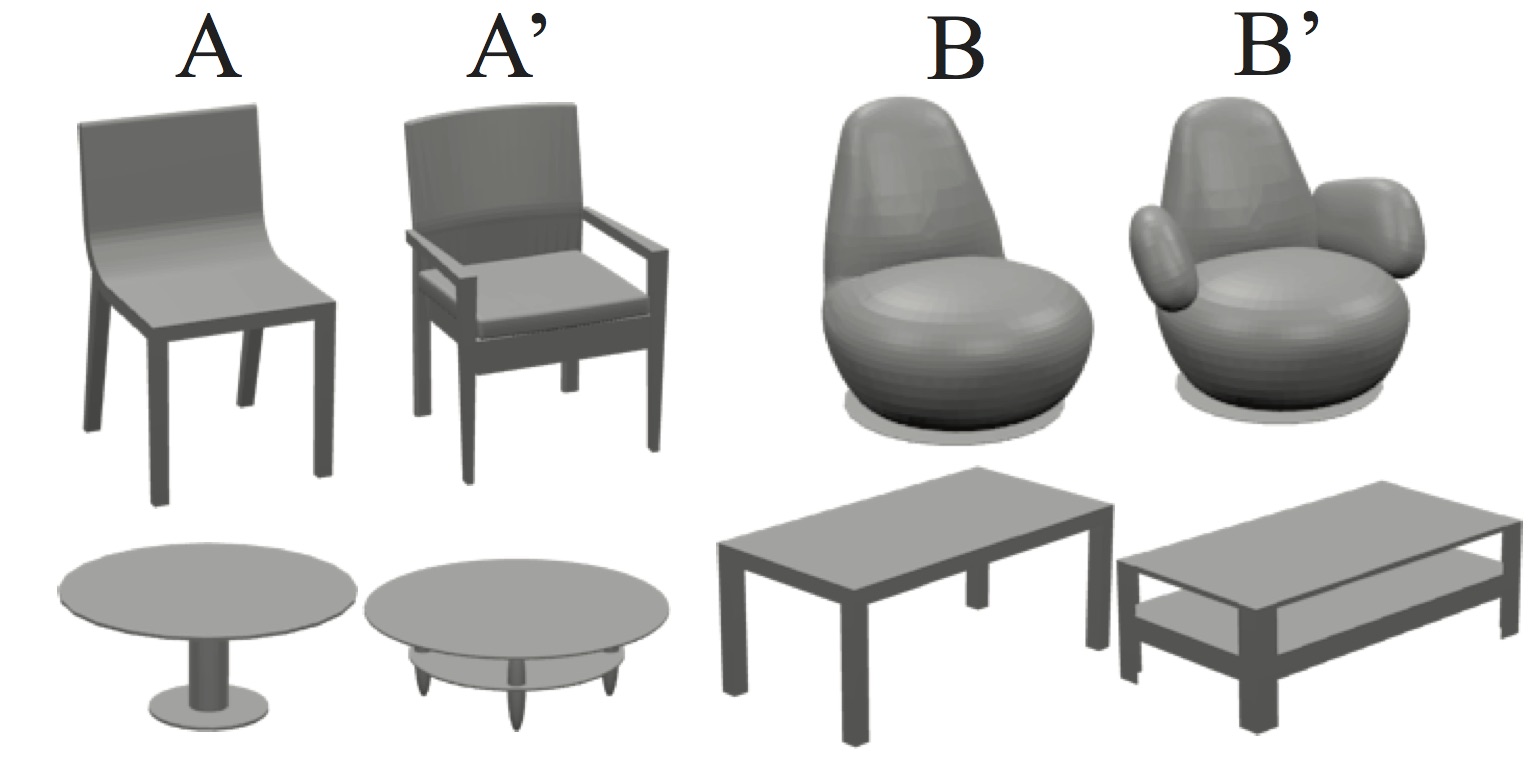}
    \vspace{-20pt}
    \caption{Shape Analogies using our learned representation. Shape $B'$ relates to $B$ in the same way that shape $A'$ relates to $A$.}
        \label{fig:shape_analogies}
\end{figure}

\begin{figure*}
    \centering
    \includegraphics[width=\textwidth]{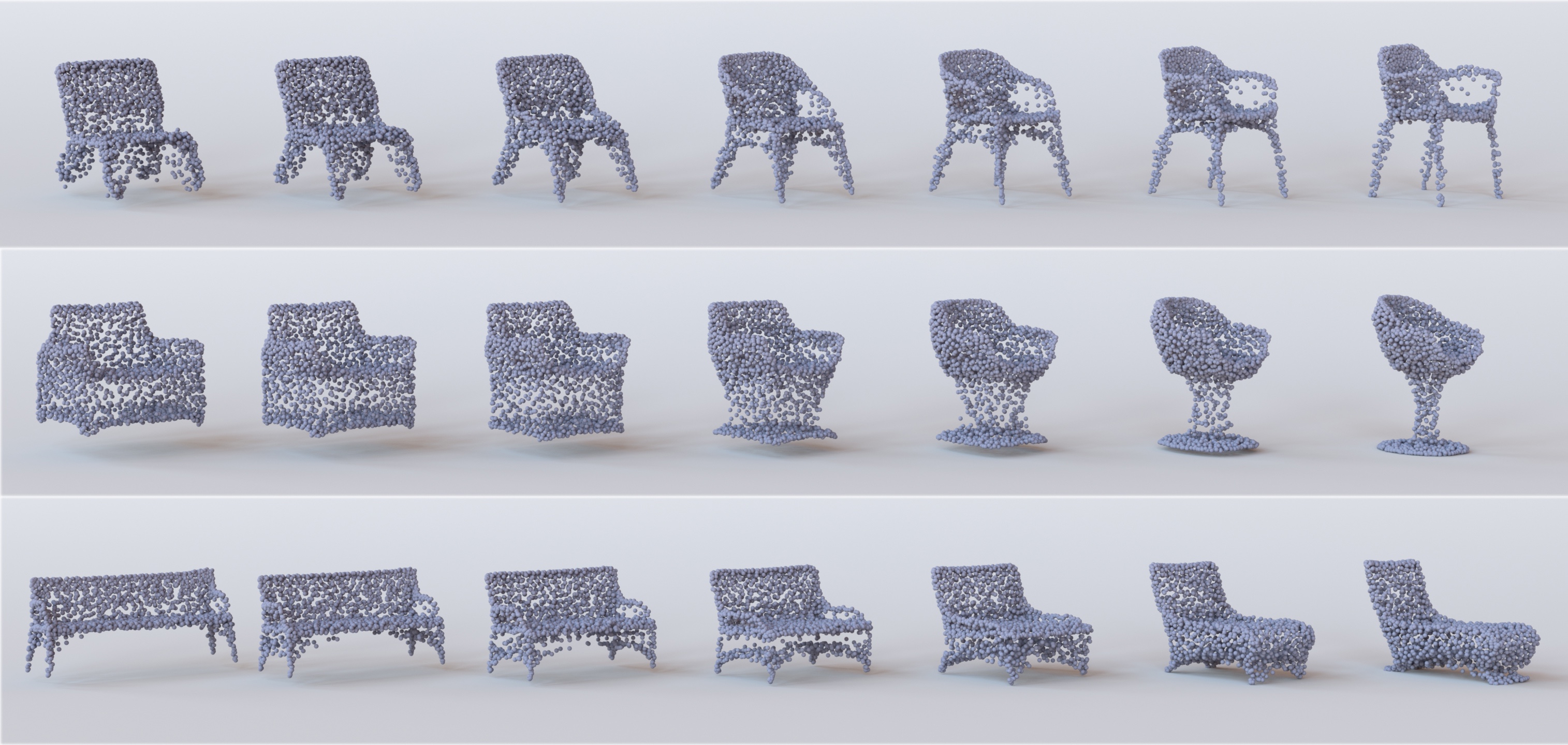}
    \vspace{-20pt}
    \caption{
    Interpolating between different point clouds (left and right-most of each row), using our latent space representation. Note the interpolation between structurally and topologically different shapes. {\bf Note}: for all our illustrations that portray point clouds we use the Mitsuba renderer~\protect\cite{Mitsuba}.}
    \label{fig:interpolations-2}
\end{figure*}

\section{Applications of the Latent Space Representation}
\label{app:applications_ae}

For shape editing applications, we use the embedding we learned with the AE-EMD trained  \emph{across all} 55 object classes, not separately per-category. This showcases its ability to encode features for different shapes, and enables interesting applications involving different kinds of shapes.

\paragraph {Editing shape parts.}
We use the shape annotations of Yi et al.\cite{yi2016_a-scalable-active-framework-for-region} as guidance to modify shapes. As an example, assume that a given object category (e.g. chairs) can be further subdivided into two sub-categories $\mathcal A$ and $\mathcal B$: every object $A \in \mathcal A$ possesses a certain structural property (e.g. has armrests, is four-legged, etc.) and objects $B\in \mathcal B$ do not. Using our latent representation we can model this structural difference between the two sub-categories by the difference between their average latent representations $\mathbf x_{\mathcal B} - \mathbf x_{\mathcal A}$, where $ \mathbf x_{\mathcal A} = \sum \limits_{A\in\mathcal A} \mathbf x_A$, $ \mathbf x_{\mathcal B} = \sum \limits_{B\in\mathcal B} \mathbf x_B $. Then, given an object $A \in \mathcal A$, we can change its property by transforming its latent representation: $x_{A'} = x_{A} + \mathbf x_{\mathcal B} - \mathbf x_{\mathcal A} $, and decode $\mathbf x_{\mathcal A'}$ to obtain $A' \in \mathcal{B}$. This process is shown in Fig.~3 of the main paper.

\paragraph{Interpolating shapes.} By linearly interpolating between the latent representations of two shapes and decoding the result we obtain intermediate variants between the two shapes. This produces a ``morph-like'' sequence with the two shapes at its end points Fig.~2 of main paper and Fig.~\ref{fig:interpolations-2} here). Our latent representation is powerful enough to support removing and merging shape parts, which enables morphing between shapes of significantly different appearance. Our cross-category latent representation enables morphing between shapes of different classes, cfg. the second row for an interpolation between a bench and a sofa.

\paragraph{Shape analogies.}
Another demonstration of the Euclidean nature of the latent space is demonstrated by finding “analogous” shapes by a combination of linear manipulations and Euclidean nearest-neighbor searching. Concretely, we find the difference vector between $A$ and $A'$, we add it to shape $B$ and search in the latent space for the nearest-neighbor of that result, which yields shape $B'$. We demonstrate the finding in Fig.~\ref{fig:shape_analogies} with images taken from the meshes used to derive the underlying point clouds to help the visualization. Finding shape analogies has been of interest recently in the geometry processing community \cite{Rustamov:2013:MEI:2461912.2461959, huang_latent}.

\begin{table}[h]
  \small
    \centering
    \begin{tabularx}{\columnwidth}{C{1} C{1} C{.8} C{1.2} C{1} C{.8} C{1.2}}
        \hline
        \multirow{2}{*}{Loss} & \multicolumn{3}{c}{ModelNet40}  & \multicolumn{3}{c}{ModelNet10}  \\
        \cmidrule(lr){2-4} \cmidrule(l){5-7}
        & $C$-plt  & icpt & loss
        & $C$-plt  & icpt & loss
        \\
        \hline
        \hline
        EMD & 0.09 & 0.5 & hng  & 0.02 & 3 & sq-hng\\
        CD  & 0.25 & 0.4 & sq-hng & 0.05 & 0.2 & sq-hng\\
        \hline
    \end{tabularx}
    \caption{Training parameters of SVMs used in each dataset with each structural loss of the AE. {\it $C$-penalty ($C$-plt)}: term controlling the trade-off between the size of the learned margin and the misclassification rate; {\it intercept (icpt)}: extra dimension appended on the input features to center them; {\it loss}: svm's optimization loss function: hinge ({\it hng}), or squared-hinge ({\it sq-hng}).}
    \label{table:svm_params}
\end{table}

\begin{figure}
    \centering
    \includegraphics[width=\columnwidth]{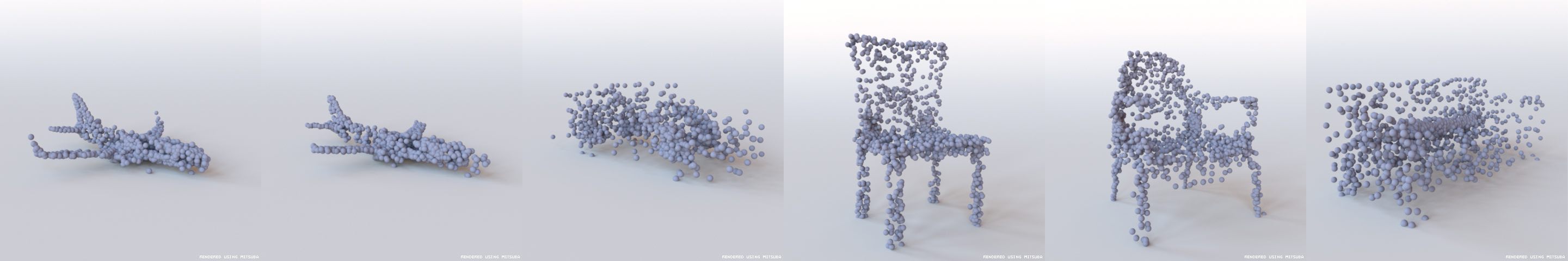}
    \vspace{-20pt}
    \caption{Synthetic results produced by the r-GAN. From left to right: airplanes, car, chairs, sofa.}
    \label{fig:generative_r_gan}
\end{figure}

\section{Autoencoding Human Forms}
In addition to ShapeNet core which contains man-made only objects, we have experimented with the D-FAUST dataset of \cite{dfaust:CVPR:2017} that contains meshes of human subjects. Specifically, D-FAUST contains $40$K scanned meshes of 10 human subjects performing a variety of motions. Each human performs a set of (maximally) $14$ motions, each captured by a temporal sequence of \tilde $300$ meshes. For our purposes, we use a random subset of $80$ (out of the $300$) meshes for each human/motion and extract from each mesh a point cloud with 4096 points. Our resulting dataset contains a total of $10240$ point clouds and we use a train-test-val split of $[70\%, 20\%, 10\%]$ - while enforcing that every split contains {\em all} human/motion combinations. We use this data to train and evaluate an AE-EMD that is identical to the single-class AE presented in the main paper, with the only difference being the number of neurons of the last layer ($4096\times 3$ instead of $2048 \times 3$).

\begin{figure}[t]
    \centering
    \includegraphics[width=\columnwidth]{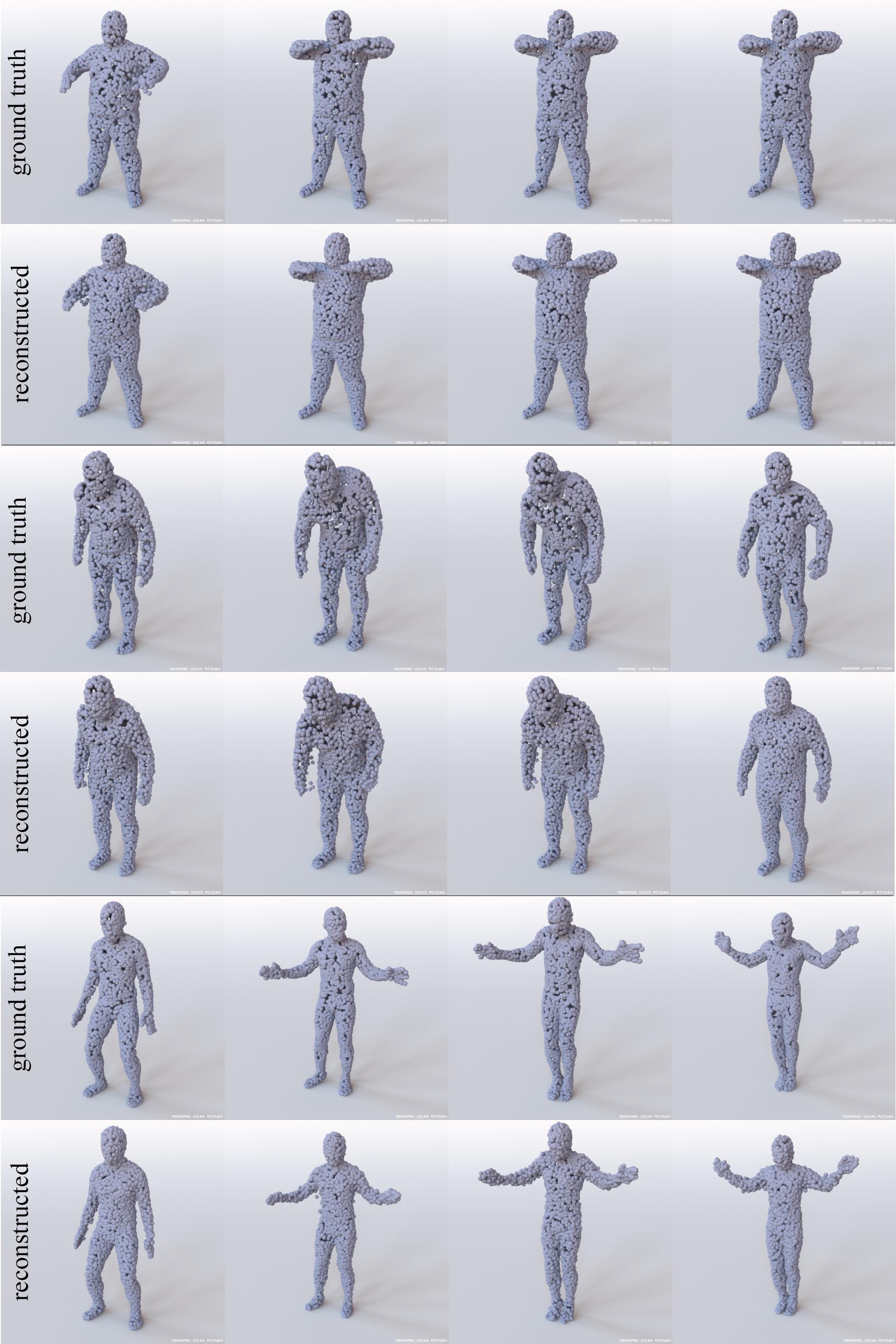}
    \vspace{-20pt}
    \caption{Reconstructions of unseen shapes from the \emph{test} split extracted from the D-FAUST dataset of \cite{dfaust:CVPR:2017} with an AE-EMD decoding point clouds with 4096 points. In each row the poses depict a motion (left-to-right) as it progress in time.}
    \label{fig:decoded_vs_gt-bogo}
\end{figure}

\begin{figure*}
    \centering
    \includegraphics[width=\textwidth]{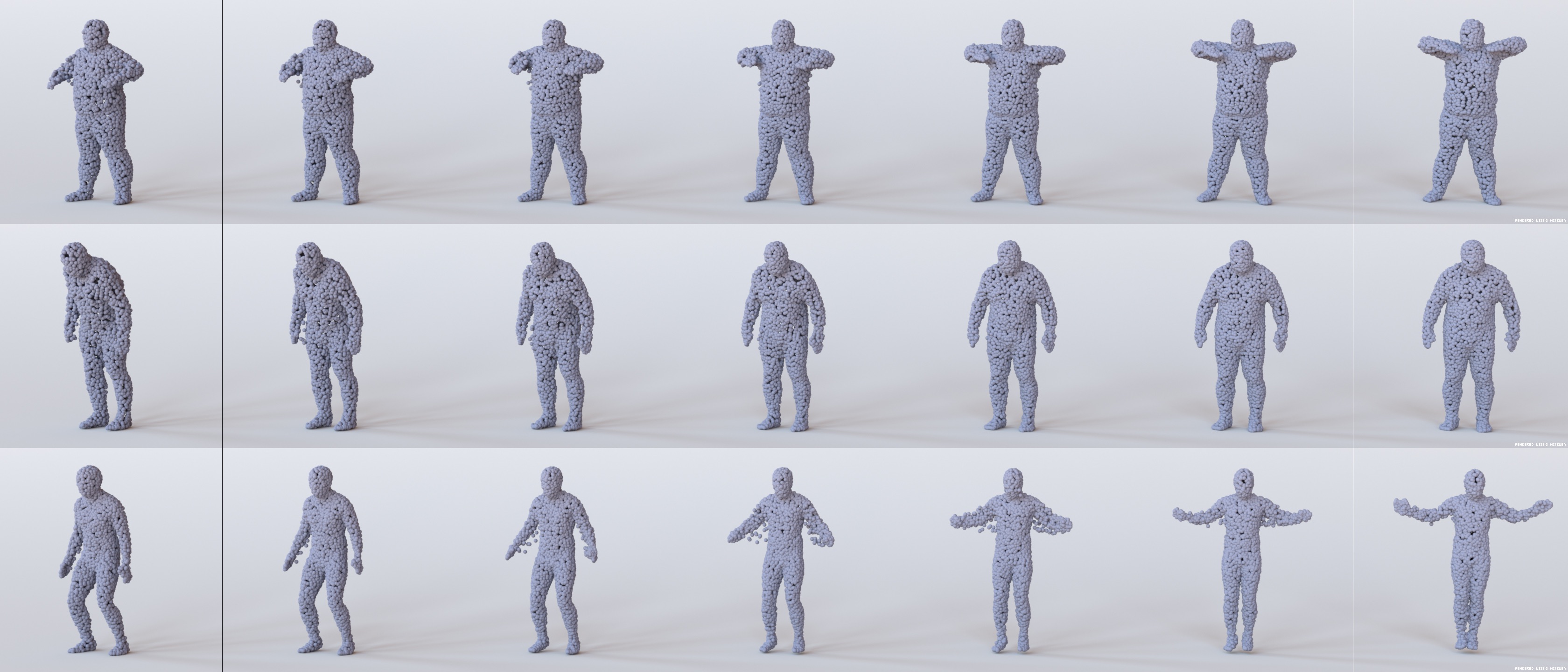}
    \vspace{-20pt}
    \caption{Interpolating between different point clouds from the {\em test} split (left and right-most of each row) of the D-FAUST dataset of \cite{dfaust:CVPR:2017}. These linear interpolations have captured some of the dynamics of the corresponding motions: 'chicken-wings' (first row), 'shake shoulders' (second row) and 'jumping jacks' (third row). Compare to Fig.\ref{fig:decoded_vs_gt-bogo} that contains ground-truth point clouds in the same time interval.}
    \label{fig:interpolations-bogo}
    \vspace{10pt}
\end{figure*}

We demonstrate reconstruction and interpolation results in Figs.~\ref{fig:decoded_vs_gt-bogo} and \ref{fig:interpolations-bogo}. For a given human subject and a specific motion we pick at random two meshes corresponding to time points $t_0$, $t_1$ (with $t_1 >  t_0$) and show their reconstructions along with the ground truth in Fig.~\ref{fig:decoded_vs_gt-bogo} (left-most and right-most of each row). In the same figure we also plot the reconstructions of two random meshes captured in $(t_0, t_1)$ (middle-two of each row). In Fig.~\ref{fig:interpolations-bogo}, instead of encoding/decoding the ground truth test data, we show decoded linear {\em interpolations} between the meshes of $t_0$, $t_1$.



\section {Shape Completions}
\label{app:shape_completions}
An important application that our AE architecture can be used for is that of completing point clouds that contain limited information of the underlying geometry. Typical range scans acquired for an object in real-time can often miss entire regions of the object due to the existence of self-occlusions and the lack of adequate (or "dense") view-point registrations. This fact makes any sensible solution to this problem of high practical importance. To address it here, we resort in a significantly different dataset than the ones used in the rest of this paper. Namely, we utilize the dataset of \cite{dai2016complete} that contains pairs of complete (intact) 3D CAD models and {\em partial} versions of them. Specifically, for each object of ShapeNet (core) it contains six partial point clouds created by the aggregation of frames taken over a limited set of view-points in a virtual trajectory established around the object. Given this data, we first fix the dimensionality of the partial point clouds to be $2048$ points for each one by randomly sub-sampling them. Second, we apply uniform-in-area sampling to each complete CAD model to extract from it $4096$ points to represent a "complete" ground-truth datum. All the resulting point clouds are centered in the unit-sphere and (within a class) the partial and complete point clouds are co-aligned. Last, we train class-specific neural-nets with Chair, Table and Airplane data and a train/val/test split of [80\%, 5\%, 15\%]. 

\subsection{Architecture}
The high level design of the architecture we use for shape-completions is identical to the AE, i.e. independent-convolutions followed by FCs, trained under a structural loss (CD or EMD). However, essential parts of this network are different: depth, bottleneck size (controlling compression ratio) and the crucial differentiation between the input and the output data. Technically, the resulting architecture is an Abstractor-Predictor (AP) and is comprised by three layers of independent per-point convolutions, with filter sizes of $[64, 128, 1024]$, followed by a max-pool, which is followed by an FC-ReLU ($1024$ neurons) and a final FC layer ($4096 \times 3$ neurons). We don't use batch-normalization between any layer and train each class-specific AP for a maximum of $100$ epochs, with ADAM, initial learning rate of $0.0005$ and a batch size of $50$. We use the minimal per the validation split model (epoch) for evaluating our models with the test data.

\subsection{Evaluation}
\begin{figure*}[t]
    \centering
    \includegraphics[width=\textwidth]{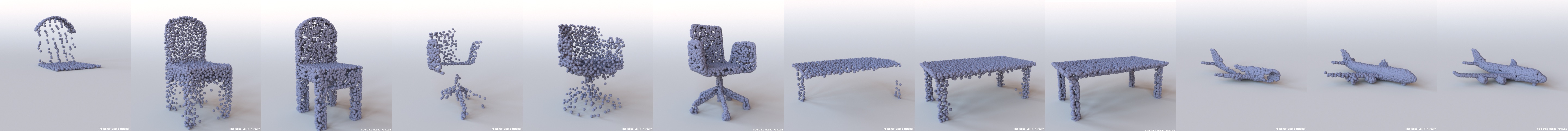}
    \vspace{-20pt}
    \caption{Point cloud {\em completions} of a network trained with partial and complete (input/output) point clouds and the {\bf CD loss}. Each triplet shows the partial input from the test split (left-most), followed by the network's output (middle) and the complete ground-truth (right-most). Also compare with Fig.~4 of main paper that portrays the corresponding completions of a network trained with the EMD loss.}
    \label{fig:completions-chamfer}
\end{figure*}

We use the specialized point cloud completion metrics introduced in \cite{sung2015data}. That is a) the {\em accuracy}: which is the fraction of the predicted points that are within a given radius ($\rho$) from any point in the ground truth point cloud and b) the {\em coverage}: which is the fraction of the ground-truth points that are within $\rho$ from any predicted point. In Table~\ref{tab:chamfer_vs_emd_on_test} we report these metrics
(with a $\rho = 0.02$ similarly to \cite{sung2015data}) for class-specific networks that were trained with the EMD and CD losses respectively. We observe that the CD loss gives rise to more accurate but also less complete outputs, compared to the EMD. This highlights again the greedy nature of CD -- since it does not take into account the matching between input/output, it can get generate completions that are more concentrated around the (incomplete) input point cloud. Figure~\ref{fig:completions-chamfer} shows the corresponding completions of those presented in the main paper, but with a network trained under the CD loss.


\begin{table}[htb!]
    \centering
    \begin{tabular}{|c||c|c|c|}
        \hline
        Class & Airplane & Chair & Table\\
        \hline
        \hline
        Test-size  & 4.5K & 6K & 6K\\
        Acc-CD      & \bf{96.9} & \bf{86.5} & \bf{87.6}\\
        Acc-EMD     & 94.7 & 77.1 & 78.4\\
        \hline
        Cov-CD      & 96.6 & 77.5 & 75.2\\
        Cov-EMD & \bf{96.8} & \bf{82.6} & \bf{83.0}\\
        \hline
    \end{tabular}
    \caption{Performance of point cloud {\em completions} on ShapeNet {\em test} data. Comparison between Abstractor-Predictors trained under the CD or EMD losses, on mean {\bf Acc}uracy and {\bf Cov}erage, across each class. The size of each test-split is depicted in the first row.}
    \label{tab:chamfer_vs_emd_on_test}
\end{table}

\section{SVM Parameters for Auto-encoder Evaluation}
\begin{figure}[ht]
    \centering
    \includegraphics[width=\columnwidth]{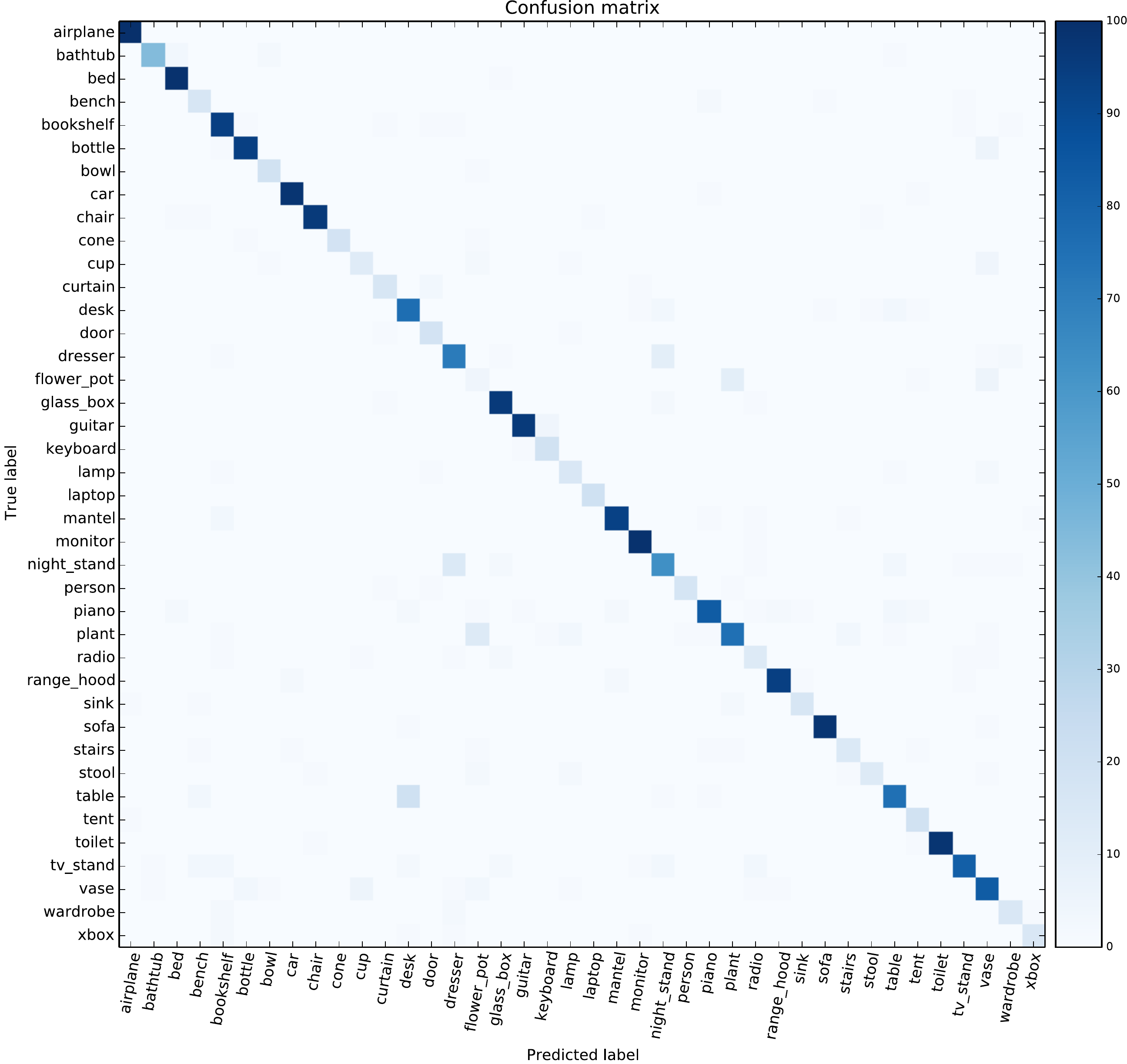}
    \vspace{-10pt}
    \caption{Confusion matrix for the SVM-based classification of Section~5.1, for the Chamfer loss on ModelNet40. The class pairs most confused by the classifier are dresser/nightstand, flower pot/plant. Better viewed in the electronic version.}
    \label{fig:confusion_matrix}
\end{figure}

For the classification experiments of Section~5.1 (main paper) we used a one-versus-rest linear SVM classifier with an $l_2$ norm penalty and balanced class weights. The exact optimization parameters can be found in Table~\ref{table:svm_params}. The confusion matrix of the classifier evaluated on our latent codes on ModelNet40 is shown in Fig.~\ref{fig:confusion_matrix}.

\section {r-GAN Details}
The discriminator's first $5$ layers are 1D-convolutions with stride/kernel of size $1$ and $\{64, 128, 256, 256, 512\}$ filters each; interleaved with leaky-ReLU. They are followed by a feature-wise max-pool. The last $2$ FC-leaky-ReLU layers have $\{128, 64\}$, neurons each and they lead to single sigmoid neuron. We used $0.2$ units of leak.

The generator consists of $5$ FC-ReLU layers with $\{64, 128, 512, 1024, 2048\times3 \}$ neurons each. We trained r-GAN with Adam with an initial learning rate of $0.0001$, and $beta_1$ of $0.5$ in batches of size $50$. The noise vector was drawn by a spherical Gaussian of $128$ dimensions with zero mean and $0.2$ units of standard deviation.

Some  synthetic results produced by the r-GAN are shown in Fig.~\ref{fig:generative_r_gan}.

\section {l-GAN Details}
The discriminator consists of $2$ FC-ReLU layers with $\{256, 512\}$ neurons each and a final FC layer with a single sigmoid neuron. The generator consists of $2$ FC-ReLUs with $\{128, k=128\}$ neurons each. When used the l-Wasserstein-GAN, we used a gradient penalty regularizer $\lambda=10$ and trained the critic for $5$ iterations for each training iteration of the generator. The training parameters (learning rate, batch size) and the generator's noise distribution were the same as those used for the r-GAN.

\section{Model Selection of GANs}
\begin{figure*}
    \centering
    \includegraphics[width=.33\linewidth]{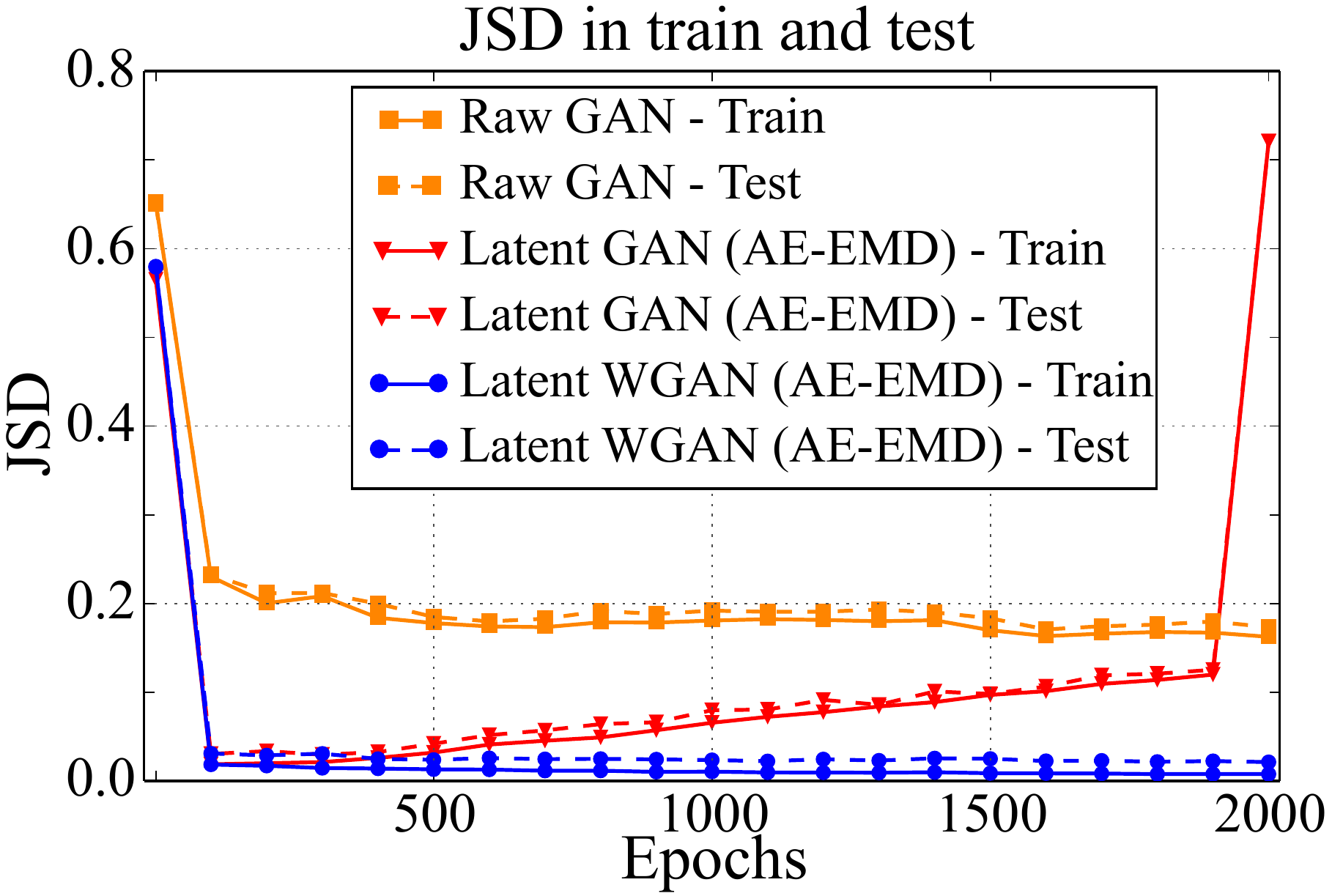}
    \includegraphics[width=.33\linewidth]{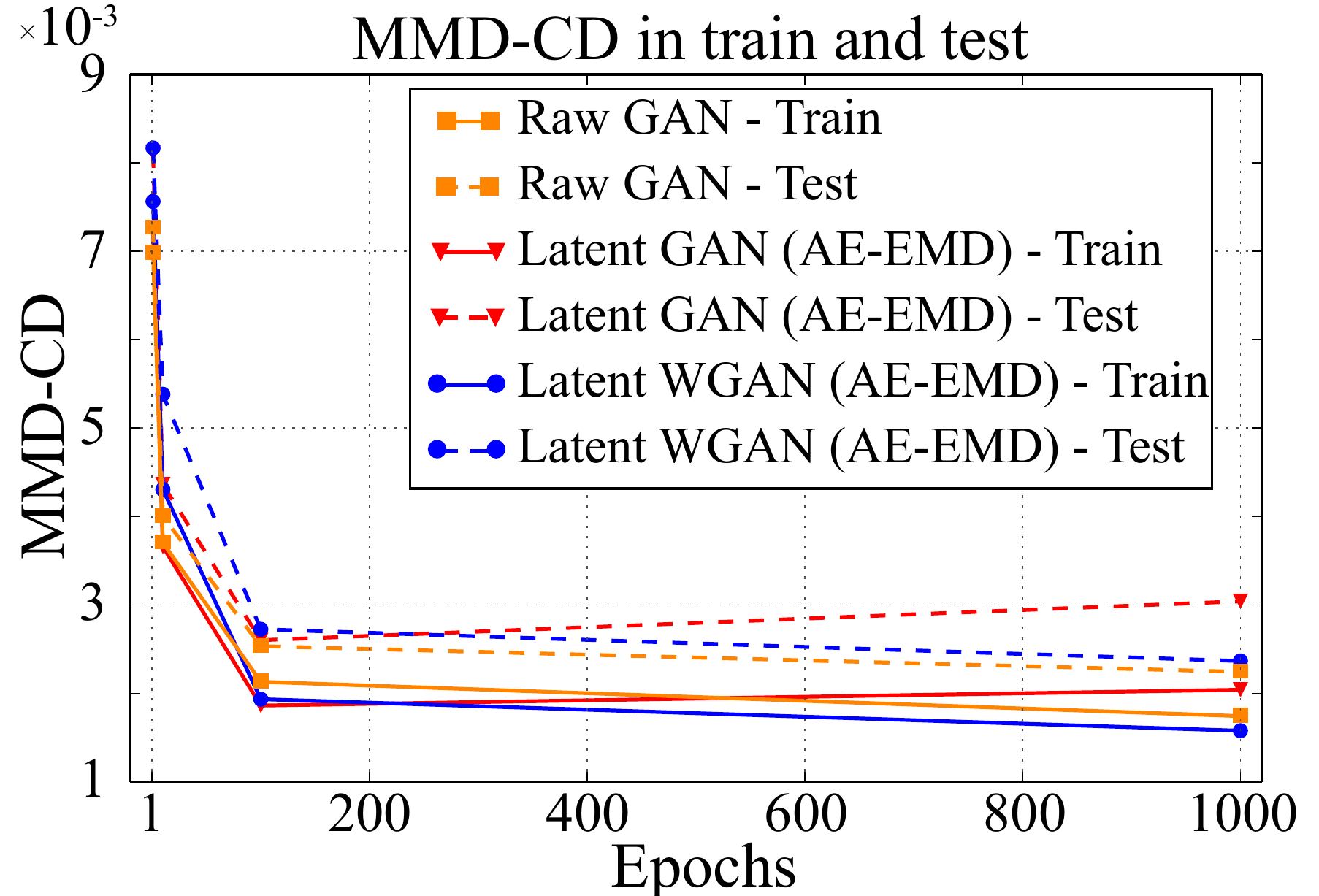}
    \includegraphics[width=.33\linewidth]{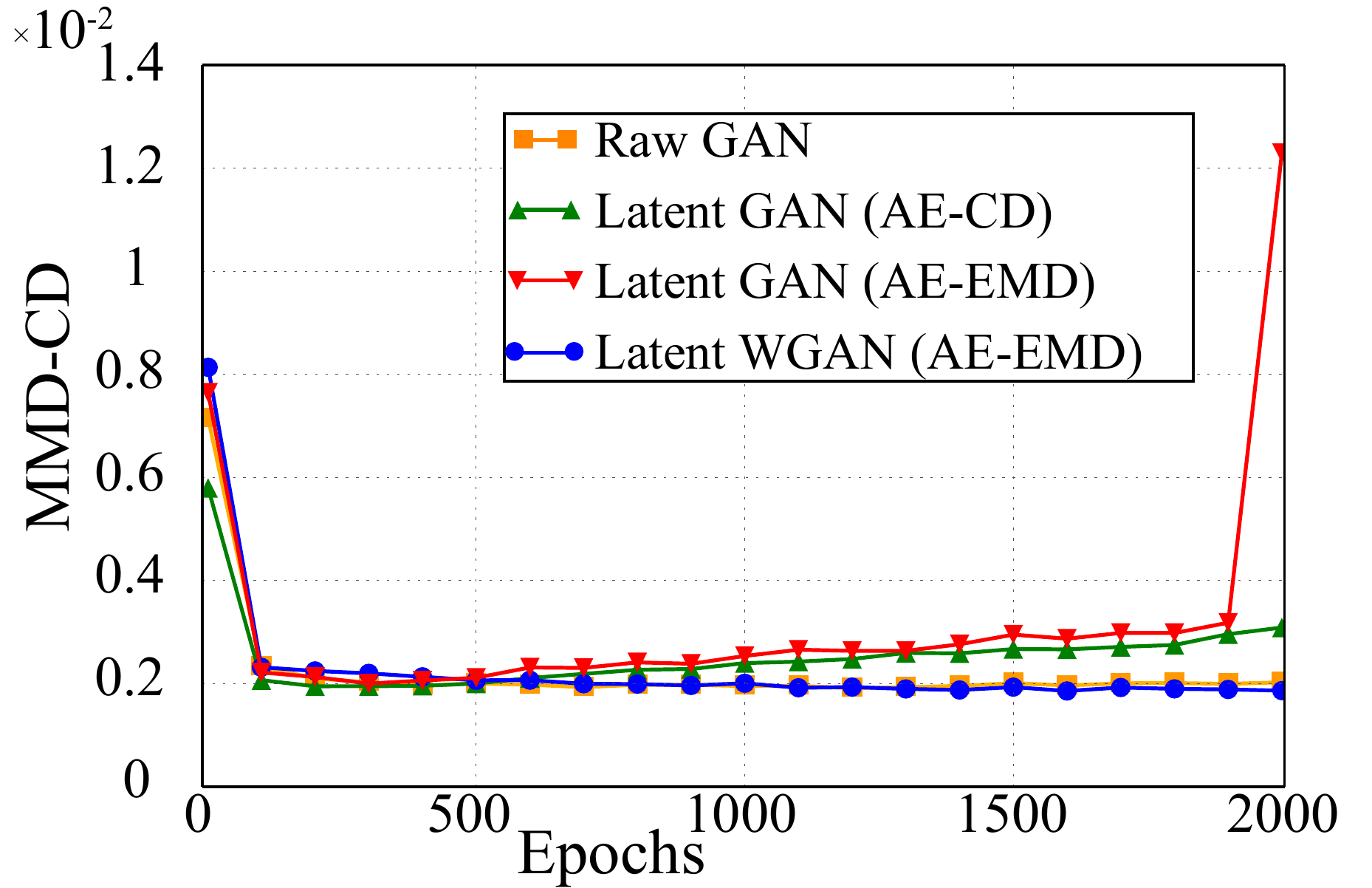}
    \vspace{-10pt}
    \caption{Left/middle: Generalization error of the various GAN models, at various training epochs. Generalization is estimated using the JSD (left) and MMD-CD (middle) metrics, which measure closeness of the synthetic results to the training resp. test ground truth distributions. The plots show the measurements of various GANs. Right: Training trends in terms of the MMD-CD metric for the various GANs. Here, we sample a set of synthetic point clouds for each model, of size 3x the size of the ground truth test dataset, and measure how well this synthetic dataset matches the ground truth in terms of MMD-CD. This plot complements Fig.~6 (left) of the main paper, where a different evaluation measure was used - note the similar behavior.}
    \label{fig:generative_generalization_model_selection_mmd}
\end{figure*}

All GANs are trained for maximally 2000 epochs; for each GAN, we select one of its training epochs to obtain the ``final'' model, based on how well the synthetic results match the ground-truth distribution. Specifically, at a given epoch, we use the GAN to generate a set of synthetic point clouds, and measure the distance between this set and the validation set. We avoid measuring this distance using MMD-EMD, given the high computational cost of EMD. Instead, we use either the JSD or MMD-CD metrics to compare the synthetic dataset to the validation dataset. To further reduce the computational cost of model selection, we only check every 100 epochs (50 for r-GAN). The generalization error of the various GAN models, at various training epochs, as measured by MMD and JSD is shown in Fig.~\ref{fig:generative_generalization_model_selection_mmd} (left and middle).

%
%

\begin{figure}
    \centering
    \begin{minipage}{\columnwidth}
  \centering
  \raisebox{-0.5\height}{\includegraphics[width=.68\columnwidth]{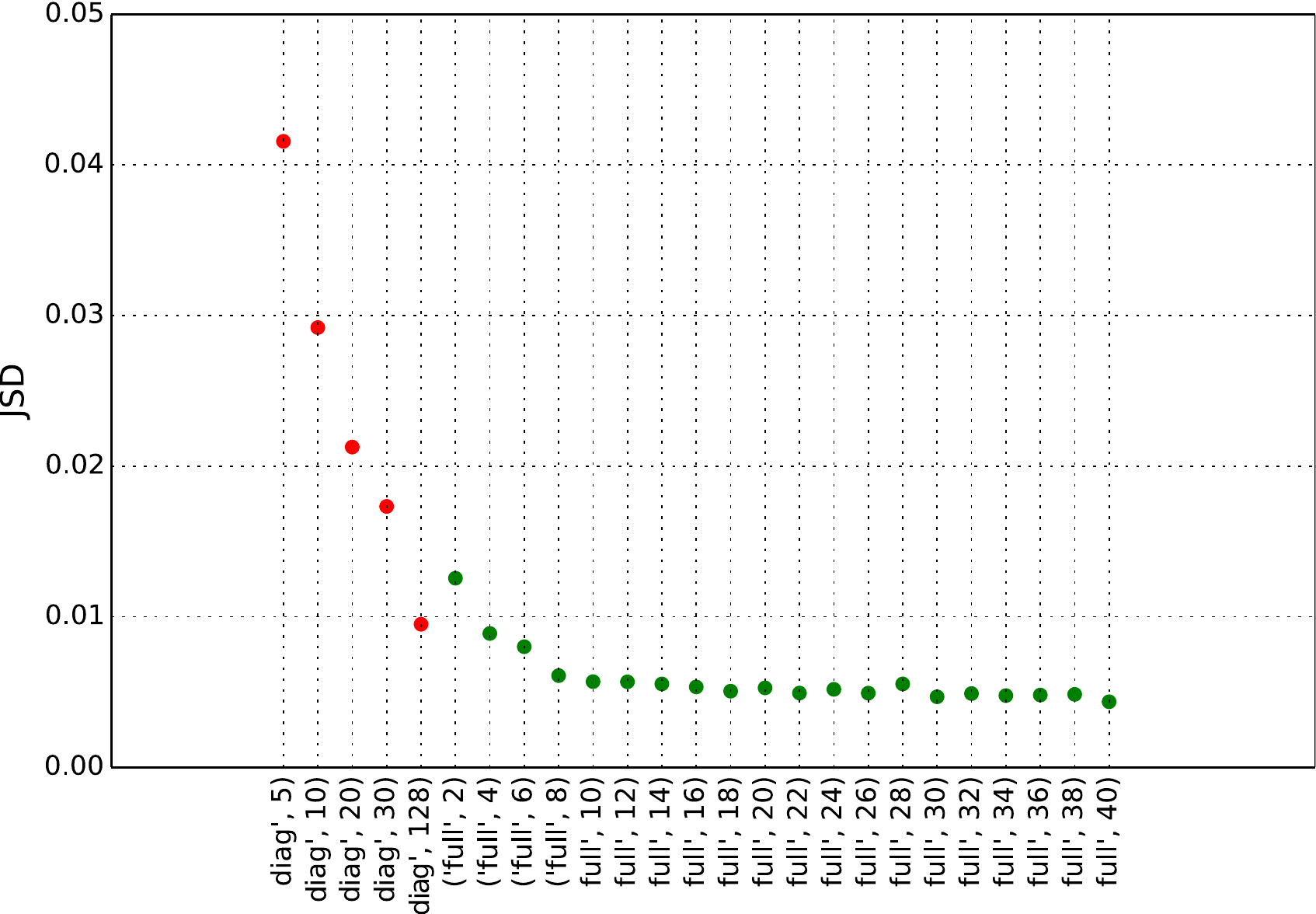}}
  \raisebox{-0.4\height}{\includegraphics[width=.3\columnwidth]{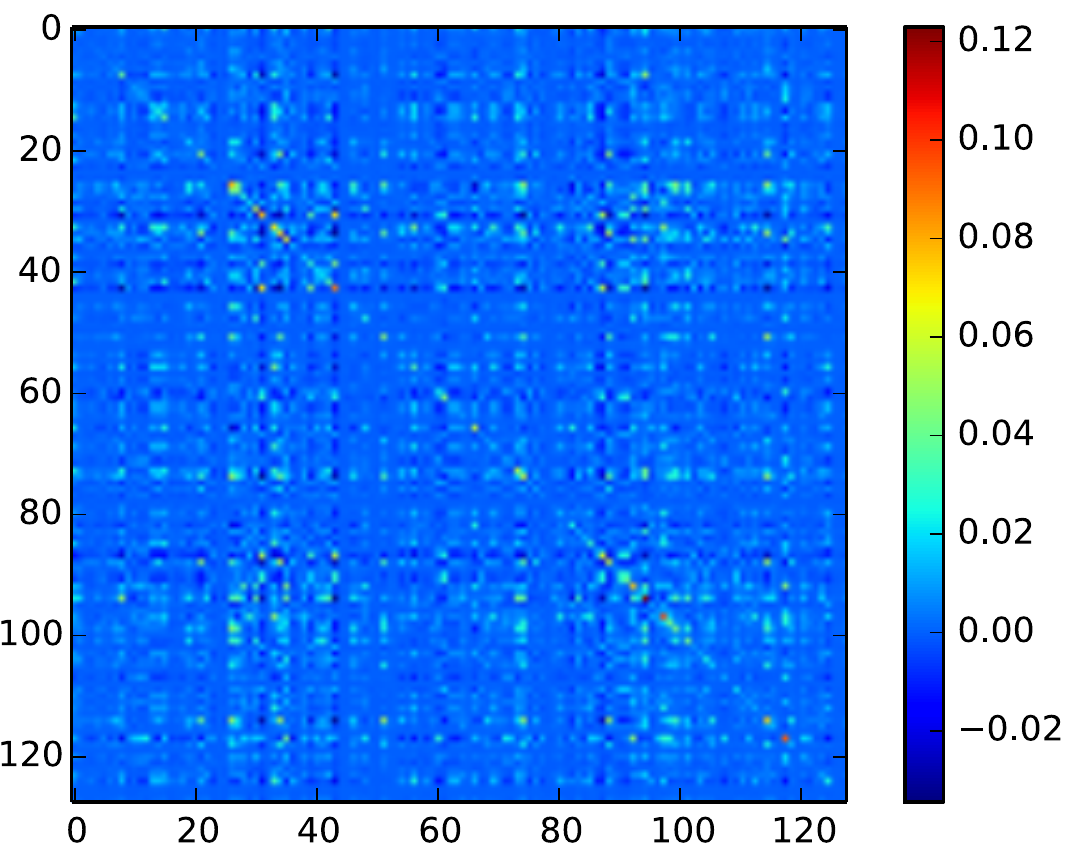}}
\end{minipage}
    \caption{GMM model selection. GMMs with a varying number of Gaussians and covariance type are trained on the latent space learned by and AE trained with EMD and a bottleneck of 128. Models with a full covariance matrix achieve significantly smaller JSD than models trained with diagonal covariance. For those with full covariance, 30 or more clusters seem sufficient to achieve minimal JSD. On the right, the values in a typical covariance matrix of a Gaussian component are shown in pseudocolor - note the strong off-diagonal components.}
    \label{fig:jsd_of_gmms_and_cov}
    \vspace{-5pt}    
\end{figure}

Using the same JSD criterion, we also select the number and covariance type of Gaussian components for the GMM (Fig.~\ref{fig:jsd_of_gmms_and_cov}, left), and obtain the optimal value of 32 components. GMMs performed much better with full (as opposed to diagonal) covariance matrices, suggesting strong correlations between the latent dimensions (Fig.~\ref{fig:jsd_of_gmms_and_cov}, right).

When using  MMD-CD as the selection criterion, we obtain models of similar quality and at similar stopping epochs (see Table~\ref{table:chair_test_data_response_selecting_by_mmd}); the optimal number of Gaussians in this case was 40. The training behavior measured using MMD-CD can be seen in Fig.~\ref{fig:generative_generalization_model_selection_mmd} (right).

%
%

\begin{table}[htb]
    \begin{center}
    \begin{tabularx}{\columnwidth}{C{1} C{1} C{1} C{1} C{1} C{1} C{1}}
        \hline
        Method & Epoch & JSD & MMD-CD & MMD-EMD  & COV-EMD & COV-CD\\
        \hline
        \hline
        \small A  & 1350 &0.1893   &0.0020  &  0.1265 & 19.4 &54.7 \\
        \small B  & 300  &0.0463   &0.0020  &  0.0800 & 32.6 &58.2 \\
        \small C  & 200  &0.0319   &0.0022  &  0.0684 & 57.6 &58.7 \\
        \small D  & 1700 &0.0240   &0.0020  &  0.0664 & 64.2 &64.7 \\
        \small E  & -    & {\bf 0.0182}  &{\bf 0.0018} & {\bf 0.0646}  & {\bf 68.6}  & {\bf 69.3} \\
        \hline
    \end{tabularx}
    \caption{Evaluation of five generators on \textbf{test-split} of {\em chair} data on epochs/models that were selected via minimal MMD-CD on the validation-split.  We report: A:~r-GAN, B:~l-GAN (AE-CD), C:~l-GAN (AE-EMD) , D:~l-WGAN (AE-EMD), E:~GMM-40-F (AE-EMD). GMM-40-F stands for a GMM with 40 Gaussian components with full covariances. The reported scores are averages of three pseudo-random repetitions. Compare this with Table~3 of the main paper. Note that the overall quality of the selected models remains the same, irrespective of the metric used for model selection.}
    \label{table:chair_test_data_response_selecting_by_mmd}
    \end{center}
\end{table}

\section{Voxel AE Details}
Our voxel-based AEs are fully-convolutional with the encoders consisting of 3D-Conv-ReLU layers and the decoders of 3D-Conv-ReLU-transpose layers. Below, we list the parameters of consecutive layers, listed left-to-right. The layer parameters are denoted in the following manner: (number of filters, filter size). Each Conv/Conv-Transpose has a stride of $2$ except the last layer of the $32^3$ decoder which has $4$. In the last layer of the decoders we do not use a non-linearity. The abbreviation "bn" stands for batch-normalization.

\begin{itemize}
  \item
  {$32^3$ - model\\
  Encoder: $\quad$ Input \textarrowright (32, 6)\textarrowright (32, 6)\textarrowright bn \textarrowright (64, 4)\textarrowright (64, 2)\textarrowright bn\textarrowright (64, 2)\\
  Decoder: $\quad$ (64, 2)\textarrowright (32, 4)\textarrowright bn\textarrowright (32, 6)\textarrowright (1, 8)\textarrowright Output
  }
  \item
  {$64^3$ - model\\
  Encoder: $\quad$ Input\textarrowright (32, 6)\textarrowright (32, 6)\textarrowright bn\textarrowright (64, 4)\textarrowright (64, 4)\textarrowright bn\textarrowright (64, 2)\textarrowright (64, 2)\\
  Decoder: $\quad$ (64, 2)\textarrowright (32, 4)\textarrowright bn\textarrowright (32, 6)\textarrowright (32, 6)\textarrowright bn\textarrowright (32, 8)\textarrowright (1, 8)\textarrowright Output
  }
\end{itemize}

We train each AE for $100$ epochs with Adam under the binary cross-entropy loss. The learning rate was $0.001$, the $\beta_1$ $0.9$ and the batch size $64$. To validate our voxel AE architectures, we compared them in terms of reconstruction quality to the state-of-the-art method of \cite{tatarchenko2017octree} and obtained comparable results, as demonstrated in Table~\ref{table:voxel_ae_iou}.
\begin{table}[ht]
    \begin{center}
    \begin{tabular}{c c c}
        \hline
        Voxel Resolution & 32 & 64\\
        \hline
        \hline
        Ours  & 92.7 &  88.4 \\
        \citep{tatarchenko2017octree} & 93.9 & 90.4\\
        \hline
    \end{tabular}
    \caption{Reconstruction quality statistics for our dense voxel-based AE and the one of \cite{tatarchenko2017octree} for the ShapeNetCars dataset. Both approaches use a $0.5$ occupancy threshold and the train-test split of \cite{tatarchenko2017octree}. Reconstruction quality is measured by measuring the intersection-over-union between the input and synthesized voxel grids, namely the ratio between the volume in the voxel grid that is 1 in both grids divided by the volume that is 1 in at least one grid. }
    \label{table:voxel_ae_iou}
    \end{center}
\end{table}

\begin{table}[ht]
    \small
    \centering
    \begin{tabularx}{\columnwidth}{C{1.8} C{.8} C{.8} C{.8} C{.8}}
        \hline
        Sample Set Size   & COV-CD & MMD-CD & COV-EMD & MMD-EMD \\
        \hline
        \hline
        Entire |Train|    & 97.3 & 0.0013 & 98.2 & 0.0545 \\
        1 $\times$ |Test| & 54.0 & 0.0023 & 51.9 & 0.0699 \\
        3 $\times$ |Test| & 79.4 & 0.0018 & 78.6 & 0.0633 \\
        \hline
        Full-GMM/32\\
        (3 $\times$ |Test|) & 68.9 & 0.0018 & 67.4 & 0.0651 \\
        \hline
    \end{tabularx}
    \caption{Quantitative results of a baseline sampling/memorizing model, for different sizes of sets sampled from the training set and evaluated against the test split. The first three rows show results of a memorizing model, while the third row corresponds to our generative model. The first row shows the results of memorizing the entire training chair dataset. The second and third rows show the averages of three repetitions of the sub-sampling procedure with different random seeds.}
    \label{table:memorization}
\end{table}

\section{Memorization Baseline}
Here we compare our GMM-generator against a model that memorizes the training data of the chair class. To do this, we either consider the entire training set or randomly sub-sample it, to create sets of different sizes. We then evaluate our metrics between these ”memorized” sets and the point clouds of test split (see Table~\ref{table:memorization}). The coverage/fidelity obtained by our generative models (last row) is slightly lower than the equivalent in size case (third row) as expected: memorizing the training set produces good coverage/fidelity with respect to the test set when they are both drawn from the same population. This speaks for the validity of our metrics. Naturally, the advantage of using a learned representation lies in learning the structure of the underlying space instead of individual samples, which enables compactly representing the data and generating novel shapes as demonstrated by our interpolations. In particular, note that while some mode collapse is present in our generative results, as indicated by the $\tilde 10\%$ drop in coverage, the  MMD of our generative models is almost identical to that of the memorization case, indicating excellent fidelity.

\section{More Comparisons with Wu et al.}

In addition to the EMD-based comparisons in Table~4 of the main paper, in Tables~\ref{table:wu_comparison}, \ref{table:cd_based_wu_comparison} we provide comparisons with \cite{wu2015_3d-shapenets:-a-deep-representation-for-volumetric} for the ShapeNet classes for which the authors have made publicly available their models. In Table~\ref{table:wu_comparison} we provide JSD-based comparisons for two of our models. In Table \ref{table:cd_based_wu_comparison} we provide Chamfer-based Fidelity/Coverage comparisons on the {\em test} split, that complement the EMD-based ones of Table 4 in the main paper.

\begin{table}[htb]
    \begin{center}
    \begin{tabularx}{\columnwidth}{c *{5}{Y}}
        \hline
        \multirow{2}{*}{Class} & A & \multicolumn{2}{c}{B} & \multicolumn{2}{c}{C}\\
        \cmidrule(lr){3-4} \cmidrule(l){5-6}
        & Tr+Te & Tr & Te & Tr & Te\\
        \hline
        \hline
        \emph{airplane} & -      & 0.0149  & 0.0268 & {\bf 0.0065} & 0.0191\\
        \emph{car}      & 0.1890 & 0.0081  & 0.0109 & {\bf 0.0063} & 0.0108\\
        \emph{rifle}    & 0.2012 & 0.0212  & 0.0364 & {\bf 0.0092} & 0.0214\\
        \emph{sofa}     & 0.1812 & {\bf 0.0102}  & 0.0102 & {\bf 0.0102} & 0.0101\\
        \emph{table}    & 0.2472 & 0.0058  & 0.0177 & {\bf 0.0035} & 0.0143\\
        \hline
    \end{tabularx}
    \caption{JSD-based comparison between A:~\cite{wu2016_learning-a-probabilistic-latent-space} and our generative models --  B:~a latent GAN, C:~our GMΜ with 32 full-covariance Gaussian components. Both B and C were  trained on the latent space of our AE with the EMD structural loss. Note that the l-GAN here uses the same ``vanilla'' adversarial objective as \cite{wu2016_learning-a-probabilistic-latent-space}. Tr: train split, Te: test split.}
    \label{table:wu_comparison}
    \end{center}
\end{table}

\begin{table}[ht]
    \begin{center}
    \begin{tabularx}{\columnwidth}{c *{4}{Y}}
        \hline
        \multirow{2}{*}{Class} & \multicolumn{2}{c}{MMD-CD} & \multicolumn{2}{c}{COV-CD}\\
        \cmidrule(lr){2-3} \cmidrule(l){4-5}
        &A & B & A & B\\
        \hline
        \hline
        \emph{airplane}  & -      &      0.0005 & -     &      71.1 \\
        \emph{car}       & 0.0015 &  \bf{0.0007}& 22.9  &  \bf{63.0}\\
        \emph{rifle}     & 0.0008 &  \bf{0.0005}& 56.7  &  \bf{71.7}\\
        \emph{sofa}      & 0.0027 &  \bf{0.0013}& 42.40 &  \bf{75.5}\\
        \emph{table}     & 0.0058 &  \bf{0.0016}& 16.7  &  \bf{71.7}\\
        \hline
    \end{tabularx}
    \caption{CD based MMD and Coverage comparison between A:~\citet{wu2016_learning-a-probabilistic-latent-space} and B:~our generative model on the {\it test} split of each class. Our generative model is a GMΜ with 32 full-covariance Gaussian components, trained on the latent space of our AE with the EMD structural loss. Note that Wu et al. used {\emph all} models of each class for training.}
    \label{table:cd_based_wu_comparison}
    \end{center}
\end{table}

\paragraph {Comparisons on training data.}
In Table~\ref{table:chair_train_data_response} we compare to \cite{wu2016_learning-a-probabilistic-latent-space} in terms of the JSD and MMD-CD on the training set of the \emph{chair} category. Since \cite{wu2016_learning-a-probabilistic-latent-space} do not use any train/test split, we perform 5 rounds of sampling 1K synthetic results from their models and report the best values of the respective evaluation metrics. We also report the average classification probability of the synthetic samples to be classified as chairs by the PointNet classifier. The r-GAN mildly outperforms \cite{wu2016_learning-a-probabilistic-latent-space} in terms of its diversity (as measured by JSD/MMD), while also creating realistic-looking results, as shown by the classification score. The l-GANs perform even better, both in terms of classification and diversity, with less training epochs. Finally, note that the PointNet classifier was trained on ModelNet, and \cite{wu2016_learning-a-probabilistic-latent-space} occasionally generates shapes that only rarely appear in ModelNet. In conjunction with their higher tendency for mode collapse, this partially accounts for their lower classification scores.
\begin{table}[htb]
  \small
  \centering
    \begin{tabularx}{\columnwidth}{C{1.9} C{.85} C{.85} C{.85} C{.85} C{.85} C{.85}}
        \hline
        Metric &
        A &
        B &
        C &
        D &
        E &
        F \\
        \hline
        \hline
        JSD             & 0.1660 & 0.1705 & 0.0372 & 0.0188 & 0.0077 & {\bf 0.0048}\\
        MMD-CD          & 0.0017 & 0.0042 & 0.0015 & 0.0018 & 0.0015 & \bf{0.0014}\\
        CLF     & 84.10  & 87.00  & 96.10  & 94.53  & 89.35  & 87.40\\
        \hline
  \end{tabularx}
    \caption{Evaluating six generators on \textbf{train-split} of {\em chair} dataset on epochs/models selected via minimal JSD on the validation-split. We report: A:~r-GAN, B:~\cite{wu2016_learning-a-probabilistic-latent-space} (a volumetric approach), C:~l-GAN(AE-CD), D:~l-GAN(AE-EMD), E:~l-WGAN(AE-EMD), F:~GMM(AE-EMD). Note that the average classification score attained by the ground-truth point clouds was 84.7\%. }
    \label{table:chair_train_data_response}
\end{table}

\section{Limitations}

\begin{figure}
    \centering
    \includegraphics[width=\columnwidth]{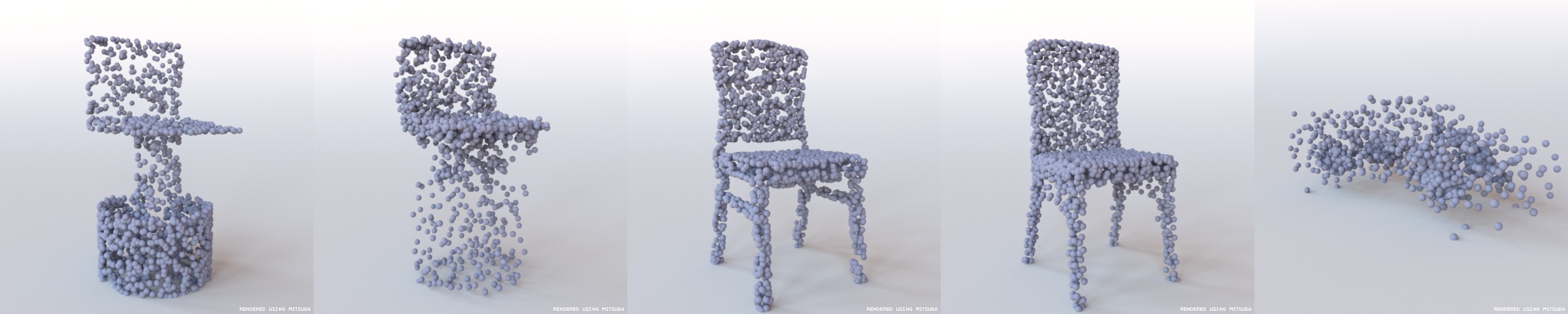}
    \vspace{-20pt}
    \caption{The AEs might fail to reconstruct uncommon geometries or might miss high-frequency details: first four images - left of each pair is the input, right the reconstruction. The r-GAN may synthesize noisy/unrealistic  results, cf. a car (right most image). }
    \label{fig:failure_cases}
\end{figure}
Figure~\ref{fig:failure_cases} shows some failure cases of our models. Chairs with rare geometries (left two images) are sometimes not faithfully decoded. Additionally, the AEs may miss high-frequency geometric details, e.g. a hole in the back of a chair (middle), thus altering the style of the input shape. Finally, the r-GAN often struggles to create realistic-looking shapes (right) -- while the r-GAN chairs are easily visually recognizable, it has a harder time on cars. Designing more robust raw-GANs for point clouds remain an interesting avenue for future work. A limitation of our shape-completion pipeline regards the style of the partial shape, which might not be well preserved in the completed point cloud (see Fig.~\ref{fig:completions-limitation} for an example).

\begin{figure*}[ht]
    \centering
    \includegraphics[width=\textwidth]{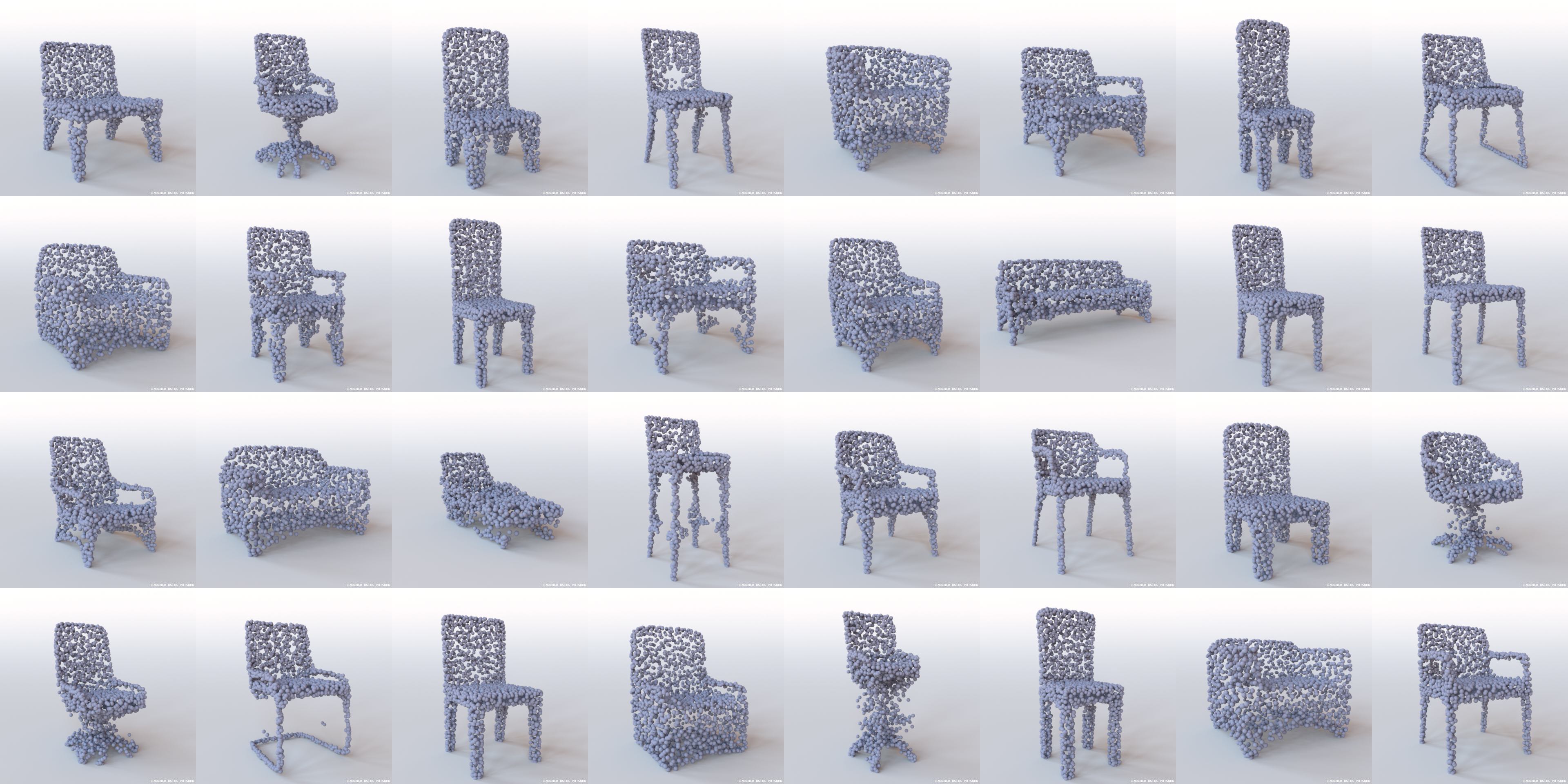}
    \caption{The 32 centers of the GMM fitted to the latent codes, and decoded using the decoder of the AE-EMD.}
    \label{fig:gmm_means_reduced_quality}
\end{figure*}

\begin{figure}[ht]
    \centering
    \includegraphics[width=\columnwidth]{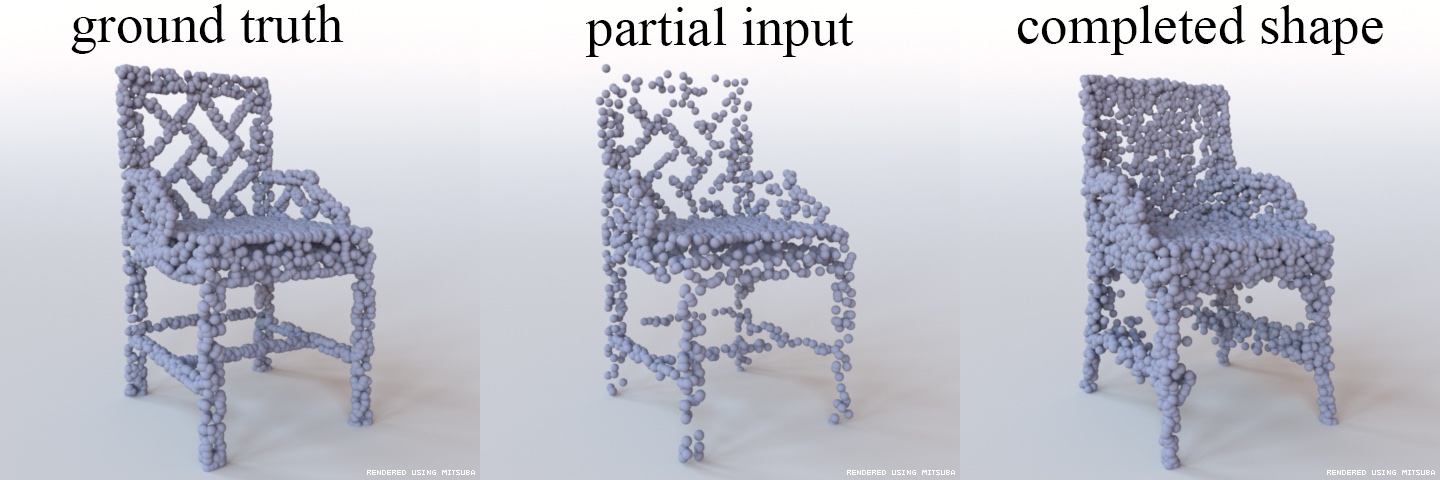}
    \caption{Our completion network might fail to preserve some of the style information in the partial point cloud, even though a reasonable shape is produced.}
    \label{fig:completions-limitation}
\end{figure}






\end{document}